%% file: main_paper.tex
\documentclass[10pt,twocolumn,letterpaper]{article}

\usepackage{cvpr}              

\usepackage{graphicx}
\usepackage{amsmath}
\usepackage{amssymb}
\usepackage{booktabs}
\usepackage{tabu}
\usepackage{multirow}
\usepackage{caption}
\usepackage{subcaption}
\usepackage{algorithm2e}
\usepackage{algpseudocode}
\usepackage{comment}
\usepackage[symbol]{footmisc}
\RestyleAlgo{ruled}
\usepackage[pagebackref,breaklinks,colorlinks]{hyperref}

\usepackage[capitalize]{cleveref}
\Crefname{section}{Section}{Sections}
\Crefname{table}{Table}{Tables}

\newcommand{\R}[0]{\mathbb{R}}
\newcommand{\x}[0]{\times}

\def\cvprPaperID{3960} 
\def\confName{CVPR}
\def\confYear{2023}

\begin{document}

\title{SimpSON: Simplifying Photo Cleanup \\ with Single-Click Distracting Object Segmentation Network}


\author{
Chuong Huynh$^{1}$\footnotemark \quad Yuqian Zhou$^{2}$ \quad Zhe Lin$^{2}$ \quad Connelly Barnes$^{2}$ \\
Eli Shechtman$^{2}$ \quad Sohrab Amirghodsi$^{2}$ \quad Abhinav Shrivastava$^{1}$ \\
$^1$University of Maryland, College Park \quad  $^2$Adobe Research \\
 $^1${\tt\small \{chuonghm,abhinav\}@cs.umd.edu}\quad $^2${\tt\small \{yuqzhou,zlin,cobarnes,elishe,tamirgho\}@adobe.com}
}

\twocolumn[{%
\renewcommand\twocolumn[1][]{#1}%
\maketitle
\vspace{-3em}
\begin{center}
    \centering
    \captionsetup{type=figure}
    \includegraphics[width=\textwidth]{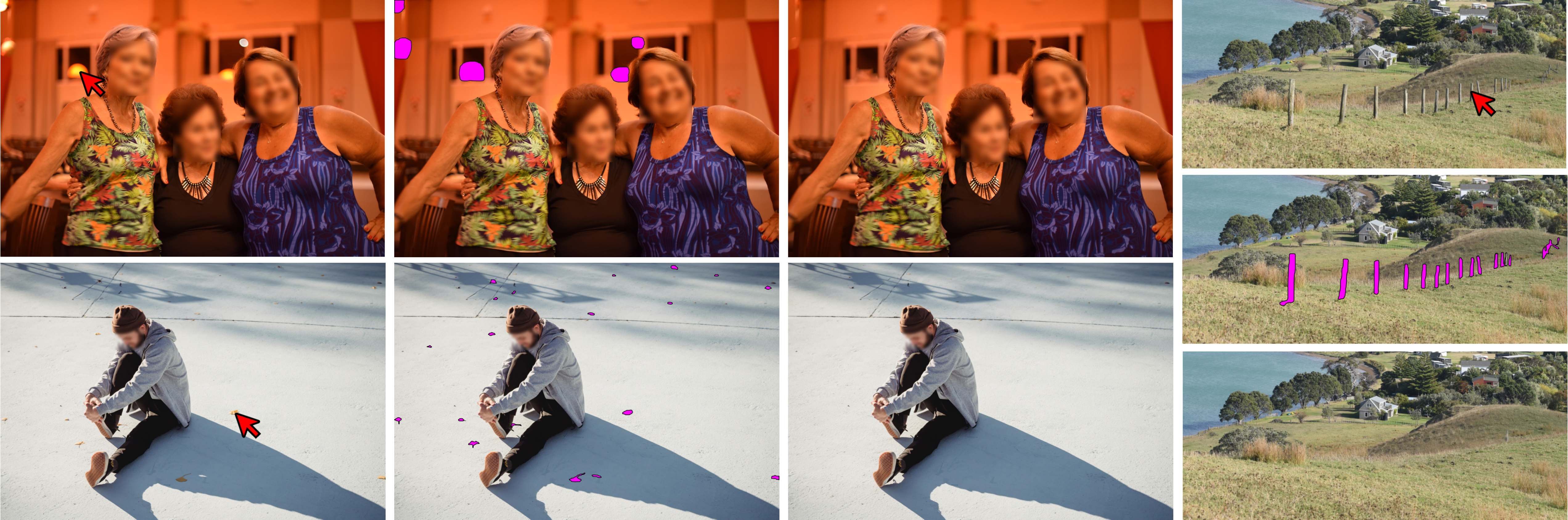}
    \captionof{figure}{
    We present a pipeline that enables the automatic segmentation of distractors in photos using a single click. With just one click, our pipeline can detect and mask the distracting object in the photo and identify other similar objects that may also be causing distraction. We can then use popular photo editing tools such as Adobe Photoshop's `Content-Aware Fill' to remove the visual distractions seamlessly. Each triad shows the input images with a click, segmentation results, and photo editing performance.}
    \label{fig:tea}
\end{center}%
}]

\input{sections/0_abstract}
\input{sections/1_introduction}
\input{sections/2_related}
\input{sections/3_methodology}
\input{sections/4_experiments}

\input{sections/5_discussion_conclusion}

{\small
\bibliographystyle{ieee_fullname}
\bibliography{main_paper}
}

\end{document}


\title{SimpSON: Simplifying Photo Cleanup \\ with Single-Click Distracting Object Segmentation Network \\ Supplementary Material}

\author{
Chuong Huynh$^{1}$ \quad Yuqian Zhou$^{2}$ \quad Zhe Lin$^{2}$ \quad Connelly Barnes$^{2}$ \\
Eli Shechtman$^{2}$ \quad Sohrab Amirghodsi$^{2}$ \quad Abhinav Shrivastava$^{1}$ \\
$^1$University of Maryland, College Park \quad  $^2$Adobe Research\\
$^1${\tt\small \{chuonghm,abhinav\}@cs.umd.edu}\quad $^2${\tt\small \{yuqzhou,zlin,cobarnes,elishe,tamirgho\}@adobe.com}
}

\twocolumn[{%
\renewcommand\twocolumn[1][]{#1}%
\maketitle
\begin{center}
    \centering
    \captionsetup{type=figure}
    \includegraphics[width=\textwidth]{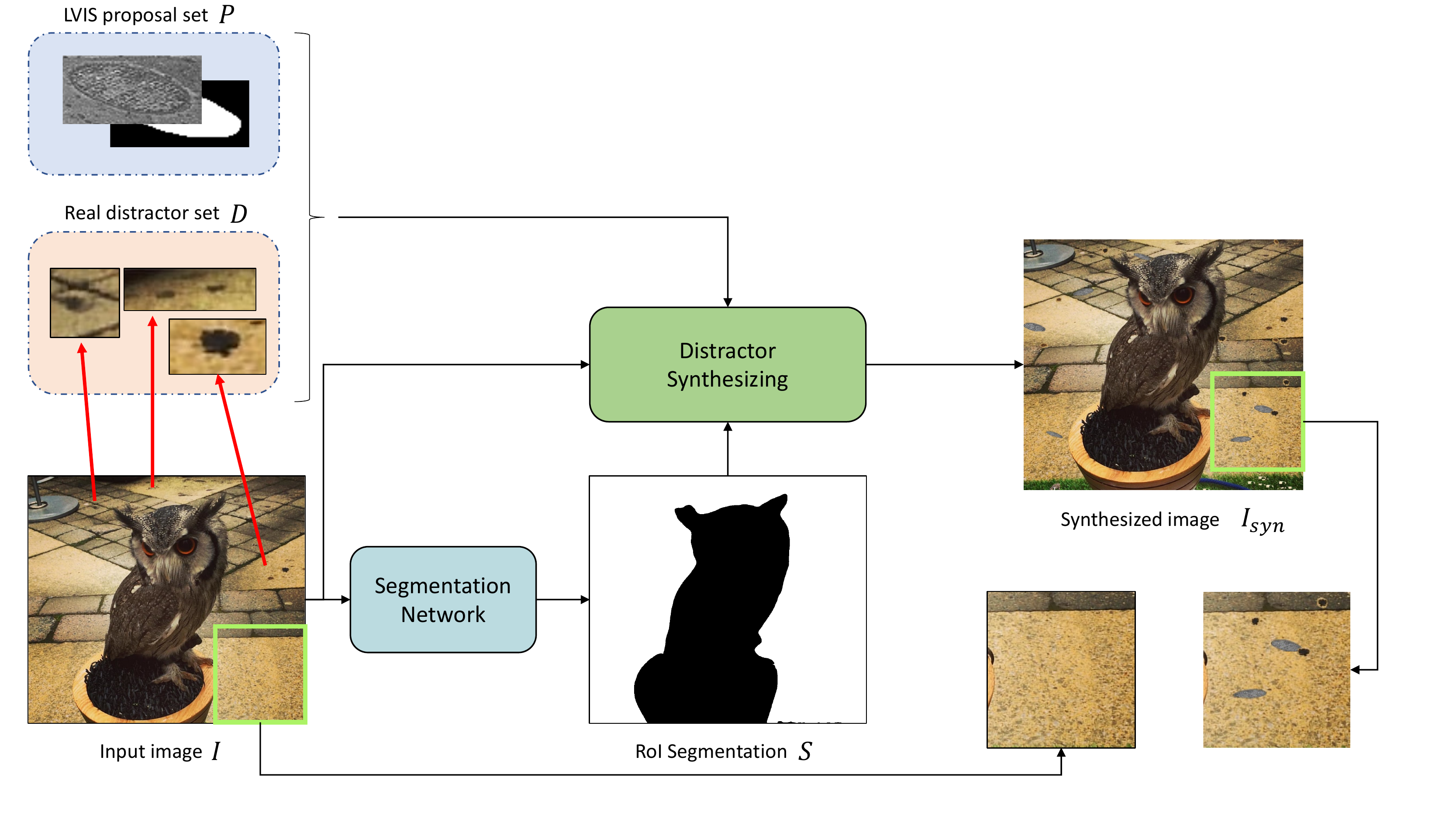}
    \captionof{figure}{The overview of our distractor synthesis pipeline. The algorithm copies real distractors and objects from LVIS database to target RoI with the help of Segmentation Network. The green box indicates the zoomed-in region before and after the synthesis pipeline. (Best view in digital color image.)}
    \label{fig:syn_pipeline}
\end{center}%
}]





\section{Distractor Synthesis Pipeline}

\setcounter{algocf}{1}
\begin{algorithm}
\caption{Distractor Synthesis Pipeline}\label{alg:data_syn}
\KwData{\\$I$: Image \\ 
$D=\{(p_i, m_i)\}$: Real sample distractor set \\
$S=\{(r_i, s_i)\}$: RoI segmentations \\
$r_i$: RoI mask \\
$s_i$ :RoI label \\
$P=\{(p_i, m_i, c_i)\}$: LVIS sample small-object set \\
$p_i$: Sample image \\
$m_i$: Sample mask \\
$c_i$: Sample RoI label}
\KwResult{$I_{syn}$, $D_{syn}$}
$D_{syn} \gets D$\;
$I_{syn} \gets I$\;
\For{$r_i, s_i$ in $S$} {
    $D' \gets$ distractors inside $r_i$\;
    $r'_i \gets r_i \setminus \{m_i | m_i \in D'\}$\;
    \uIf{$|D'| < 3$}{
        $P' \gets$ random from $P$ s.t. $c_i = s_i$\;
        $D' \gets D' \cup P'$\;
    }
    $\bar{D'} \gets$ total area of distractors in $D'$ \;
    $n \gets \lceil 10\% \times area(r'_i) / \bar{D'} \rceil$ \;
    \For{$d$ in $D'$}{
        $p_j \gets$ image crop in $d$ (real or LVIS)\;
        $m_j \gets$ mask crop in $d$ (real or LVIS)\;
        \For{$k$ in 1..$n$}{
            $\delta_i \gets$ distance map of $r'_i$\;
            $x, y \gets $ center coordinate in $\delta_i$\;
            $p'_j, m'_j \gets $augment $p_j, m_j$ \;
            Move $p'_j, m'_j$ to $(x,y)$\;
            $I'_{syn} \gets$ blend $p'_j$ to $I_{syn}$\;
            \uIf {$\|hist(I'_{syn}(x, y)) - hist(I_{syn}(x,y))\| > 0.001$} {
                $I_{syn} \gets I'_{syn}$\;
                $D_{syn} \gets D_{syn} \cup \{(p'_j, m'_j)\}$\;
                $r'_i \gets r_i \setminus m'_j $\;
            }
        }
        
    }
}
\end{algorithm}

We present the algorithm of our dataset synthesis procedure in the Algorithm \ref{alg:data_syn}. The ``DistractorSyn14k" and ``DistractorSyn-Val" datasets are obtained as described in the main paper. For each Region of Interest (RoI), which refers to the segmentation stuff regions where we intent to copy the distractors to, we sample a set of distractor objects from our self-collected real distractor dataset, and also normal objects from LVIS database. We name them ``distractor samples". The number of copies $n$ for each sample is constrained by ensuring the total area of the pasted distractors is not exceeding 10\% of the RoI. We apply random spatial and color augmentation including random flip, scale, rotate, brightness and contrast adjustment to those distractor samples before copying and pasting them. Gaussian smoothing is also applied while blending them with the image to avoid sharp composition boundary artifacts. It will avoid the model from overfitting the seams. The positions of the distractor placement can be decided by the distance map peak value, and the details are shown in the Algorithm \ref{alg:data_syn}.


\begin{figure*}[t]
    \centering
    \includegraphics[width=\textwidth]{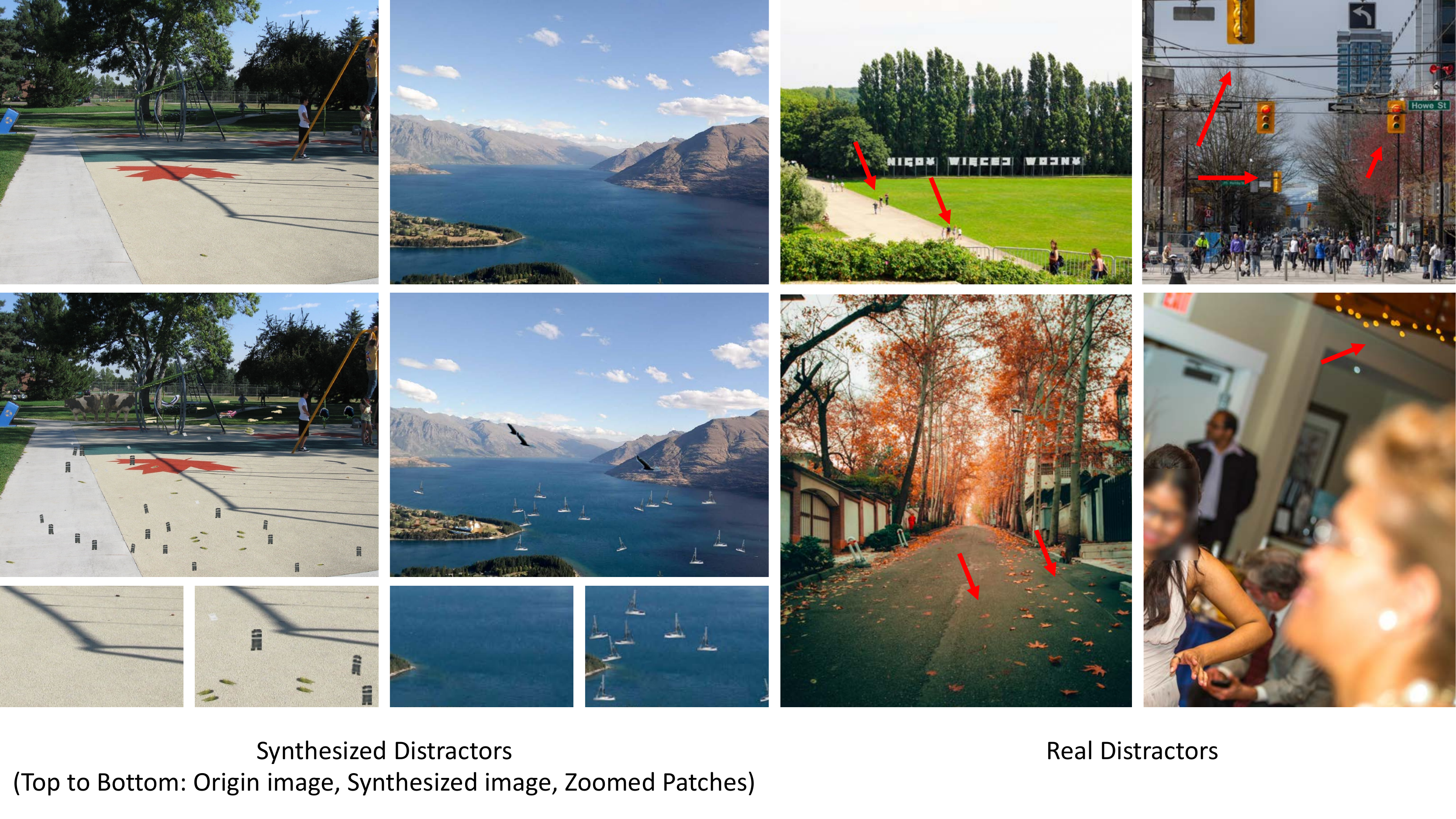}
    \caption{Examples of synthesized and real distractors. Similar to real photos, our synthesized database contains distractors with different appearances and categories. (Best view in digital color image.)}
    \label{fig:syn_examples}
\end{figure*}

\Fref{fig:syn_pipeline} illustrates the entire process and intermediate results of our data synthesis pipeline. We tend to make the synthesized images look realistic, though the compositional artifacts still exist. The resulting images still look natural enough since the distractor samples are from real distractor datasets and real small objects. According to our experiments, using synthetic data will not greatly influence the generalization ability of the model to real images. More image harmonization and deep composition techniques can be further explored in the future work. Some other results are shown in \Fref{fig:syn_examples}. Our synthetic images are comparable with real distractor ones which can be used to train and evaluate the CPN module. Our images contains many repeated distractors with diversity in appearances and categories, simulating the properties of distractors in the real-world.



\section{More details about the Distractor20K dataset}

\begin{figure}
    \centering
    \includegraphics[width=\columnwidth]{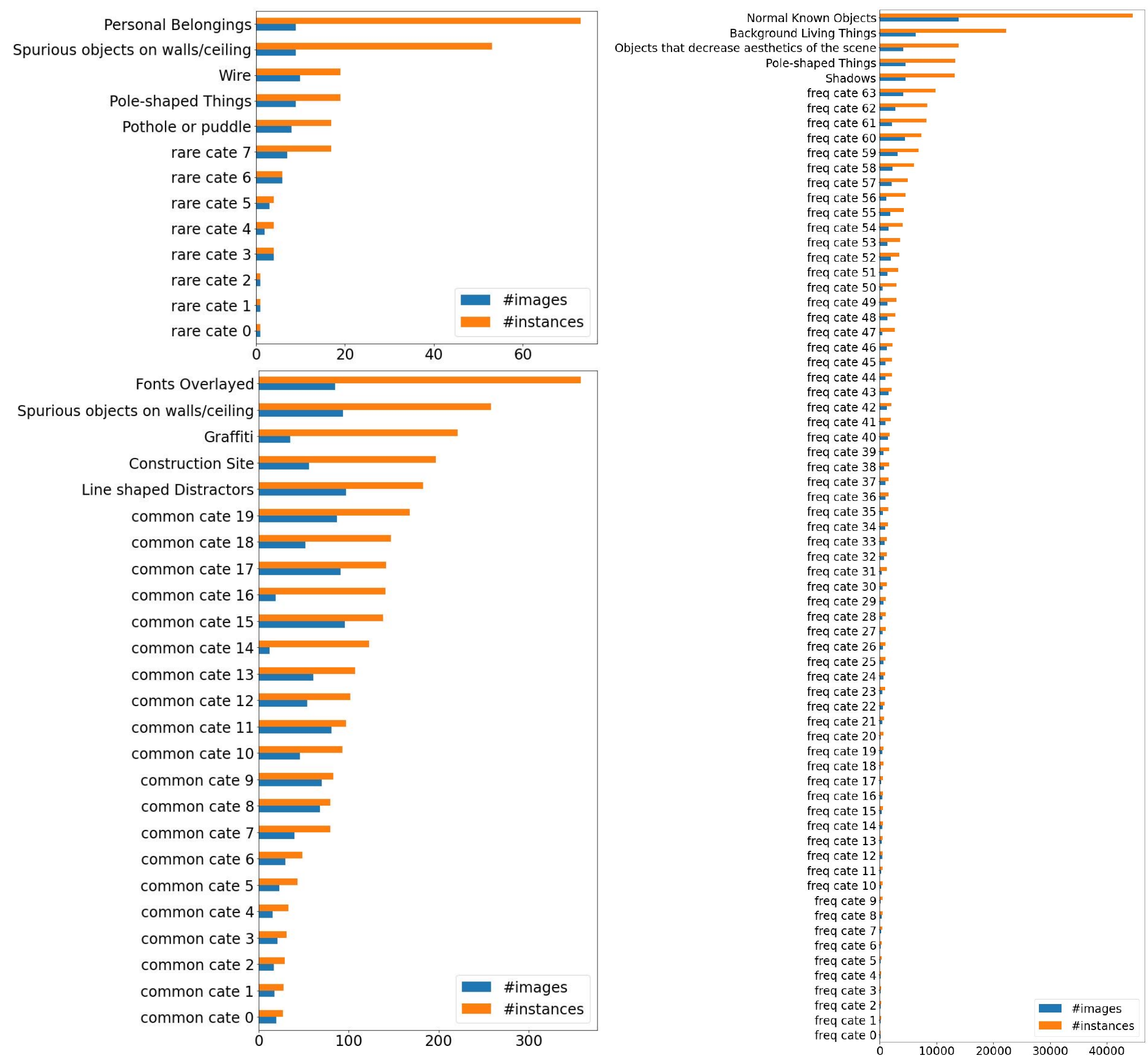}
    \caption{Frequencies of categories in the Distractor20K}
    \label{fig:data_freq}
\end{figure}

Our dataset Distractor20K is collected to have 107 different categories belonging to 28 super categories. There exist known and unknown categories that label unrecognizable regions or not in the defined category set. Both stuff and things are considered distractors in our dataset; in detail, there are 79 object categories and 28 non-object categories. If we follow the LVIS dataset to split the categories based on their frequencies, there are 13 rare, 25 common, and 69 frequent categories in our collected dataset. The number of instances and images for each category can be seen in \Fref{fig:data_freq}. Rare categories appear in a maximum of ten images in the entire dataset, while common categories have less than 100 images. All the category names are hidden for commercial use.

\begin{figure}
    \centering
    \includegraphics[width=0.49\columnwidth]{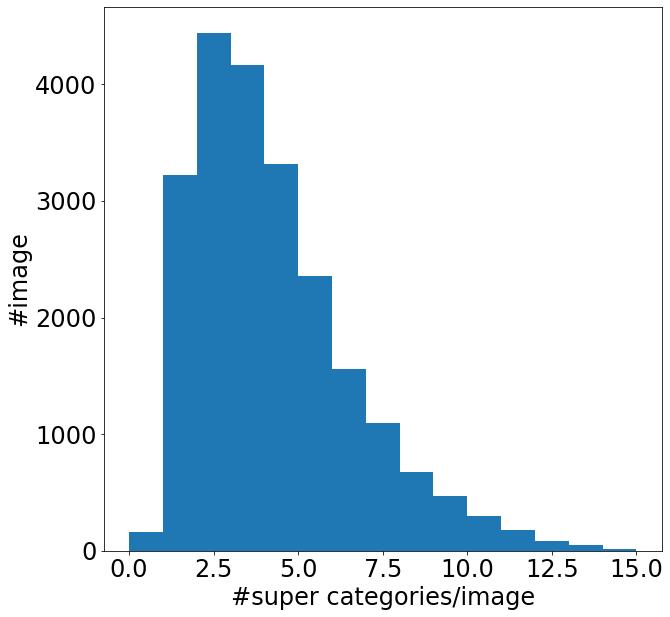}
    \includegraphics[width=0.49\columnwidth]{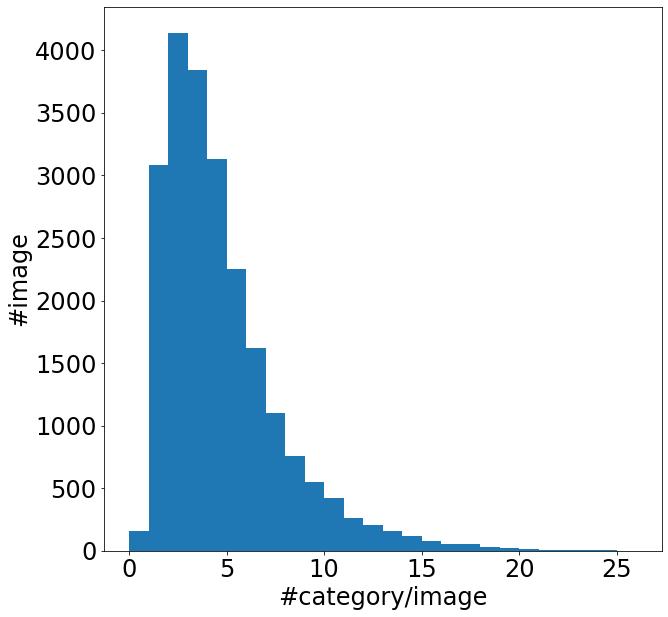}
    \caption{The number of categories in each image in the Distractor20K}
    \label{fig:data_detail}
\end{figure}

The \Fref{fig:data_detail} illustrates the histogram of a number of distractor categories for each image. An image can have up to 25 categories of 15 super categories, and the average contains 3-5 categories.

\begin{figure}[t]
    \centering
    \includegraphics[width=0.5\columnwidth]{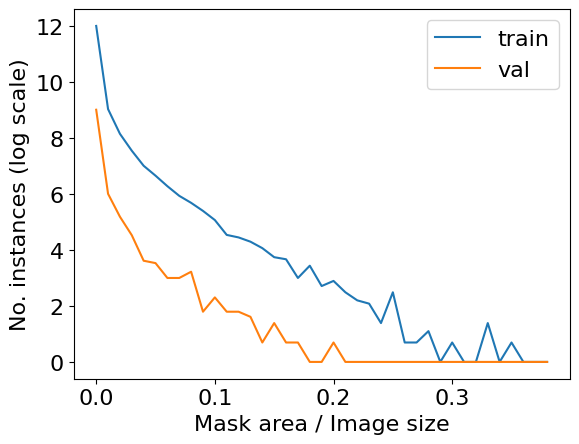}
    \caption{The number of instances regards to the object size in our distractor datasets.}
    \label{fig:object_size}
\end{figure}

Figure \ref{fig:object_size} shows the histogram of the ratio of the instance mask size over the image size in our Distractor20K. According to the statistics, we found that a significant amount of distracting instances are medium and small, as tiny as only occupying 0.01 of the image. Those distractors can be stones on the ground, graffiti on the wall, leaves on the water's surface, fire valves on the ceiling, etc. Photographers have the requests to clean those things from the photos, while existing segmentation models do not help segment them automatically. 

\section{More Results of 1C-DSN}
\subsection{Existence of Negative Clicks}

Additional experiments are executed with FocalClick and RiTM on the LVIS dataset. To validate the capability of those models with the one-click procedure, we finetuned the models on LVIS dataset with only one positive click as input. In testing, the click generator is customized to produce positive clicks only. No additional clicks are added when there are only a few false negatives at the boundary because clicking at the boundary can cause severe accuracy degrading due to the precision of click positions.  

\begin{figure}[t]
    \centering
    \includegraphics[width=\columnwidth]{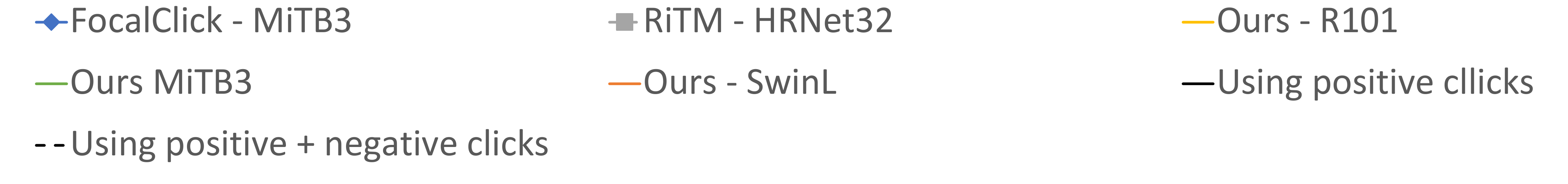}
    \begin{subfigure}[b]{0.49\columnwidth}
        \includegraphics[width=\textwidth]{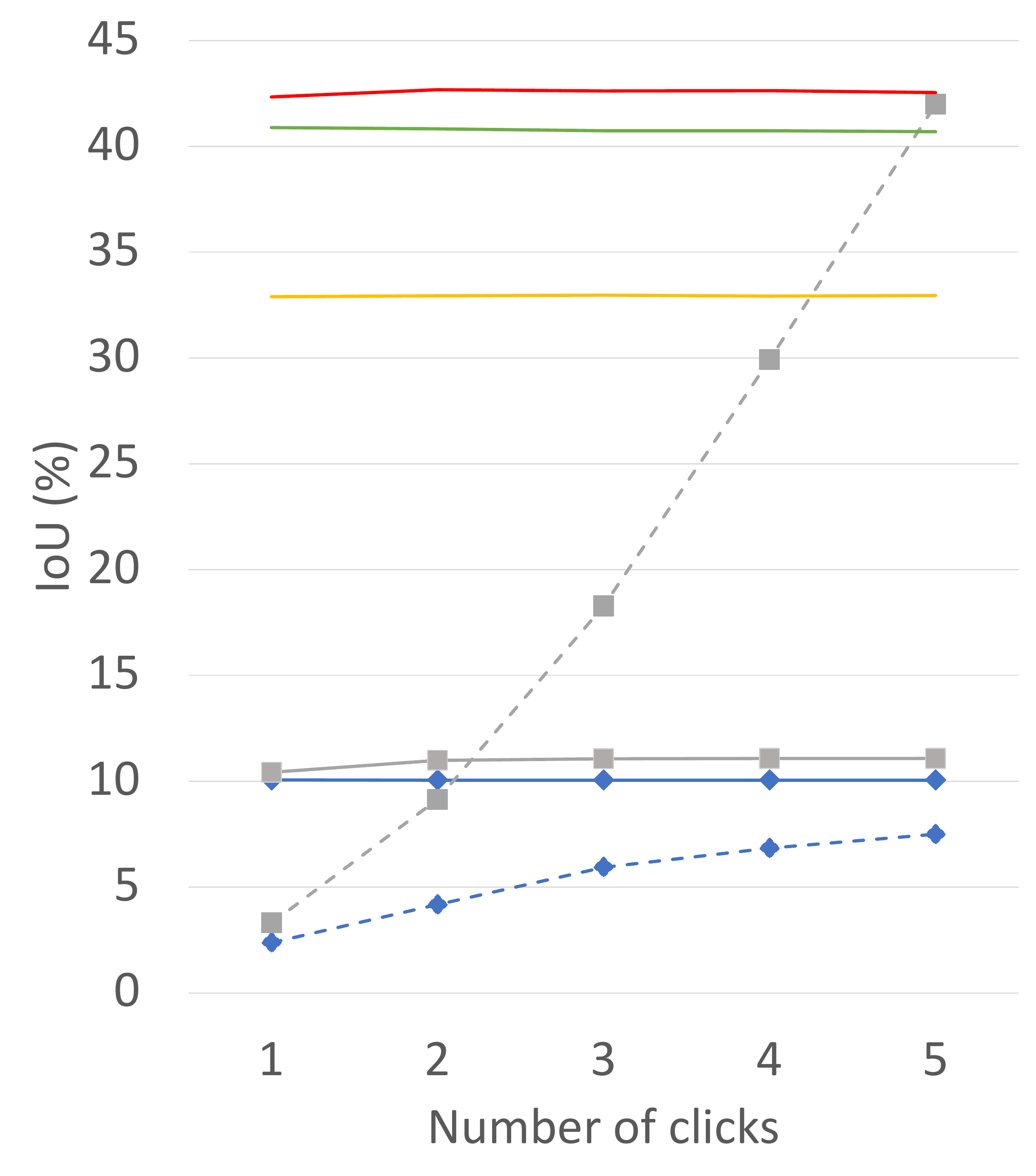}
        \caption{Small (LVIS val)}
    \end{subfigure}
    \begin{subfigure}[b]{0.49\columnwidth}
        \includegraphics[width=\textwidth]{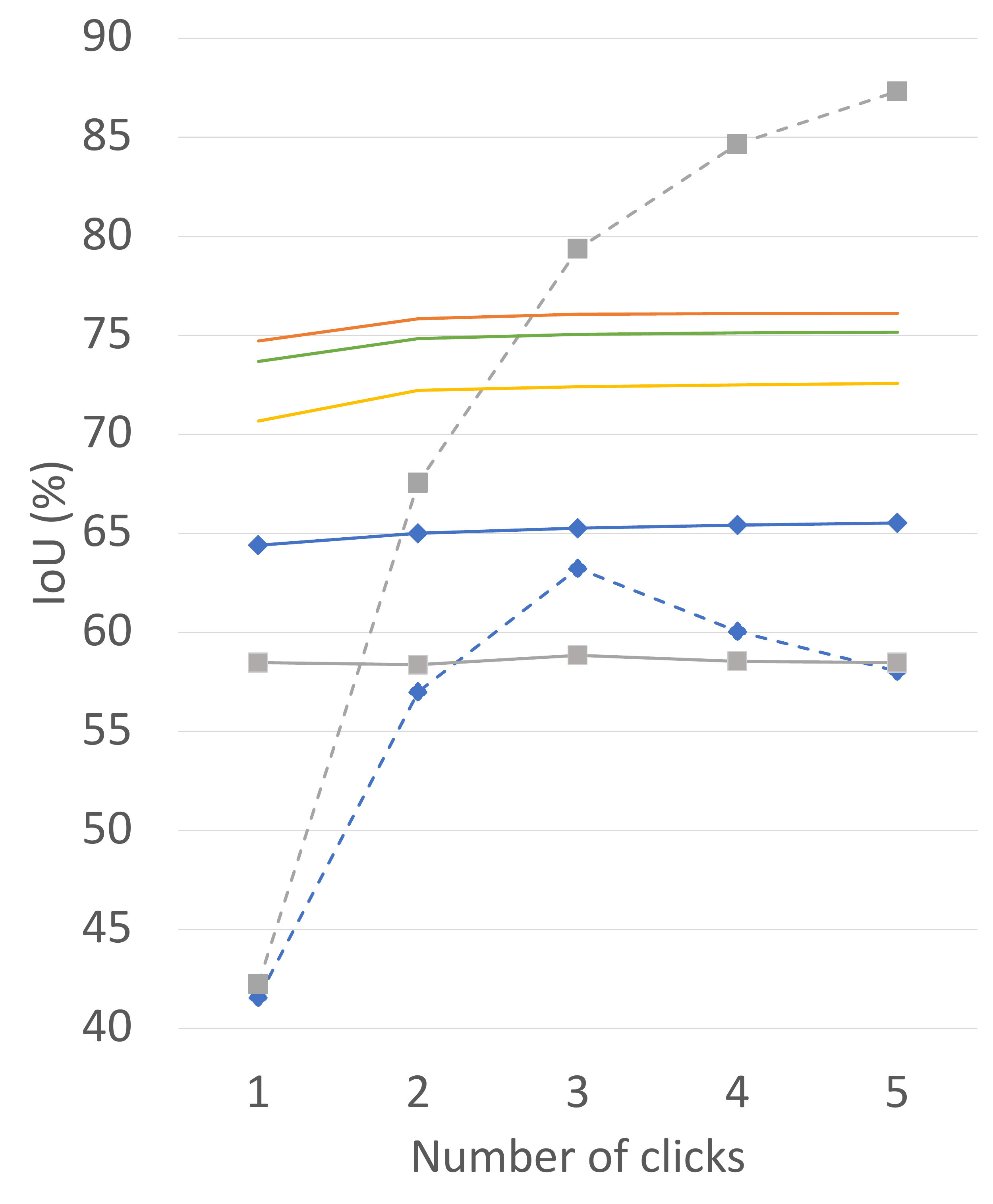}
        \caption{Medium (LVIS val)}
    \end{subfigure}
    \begin{subfigure}[b]{0.49\columnwidth}
        \includegraphics[width=\textwidth]{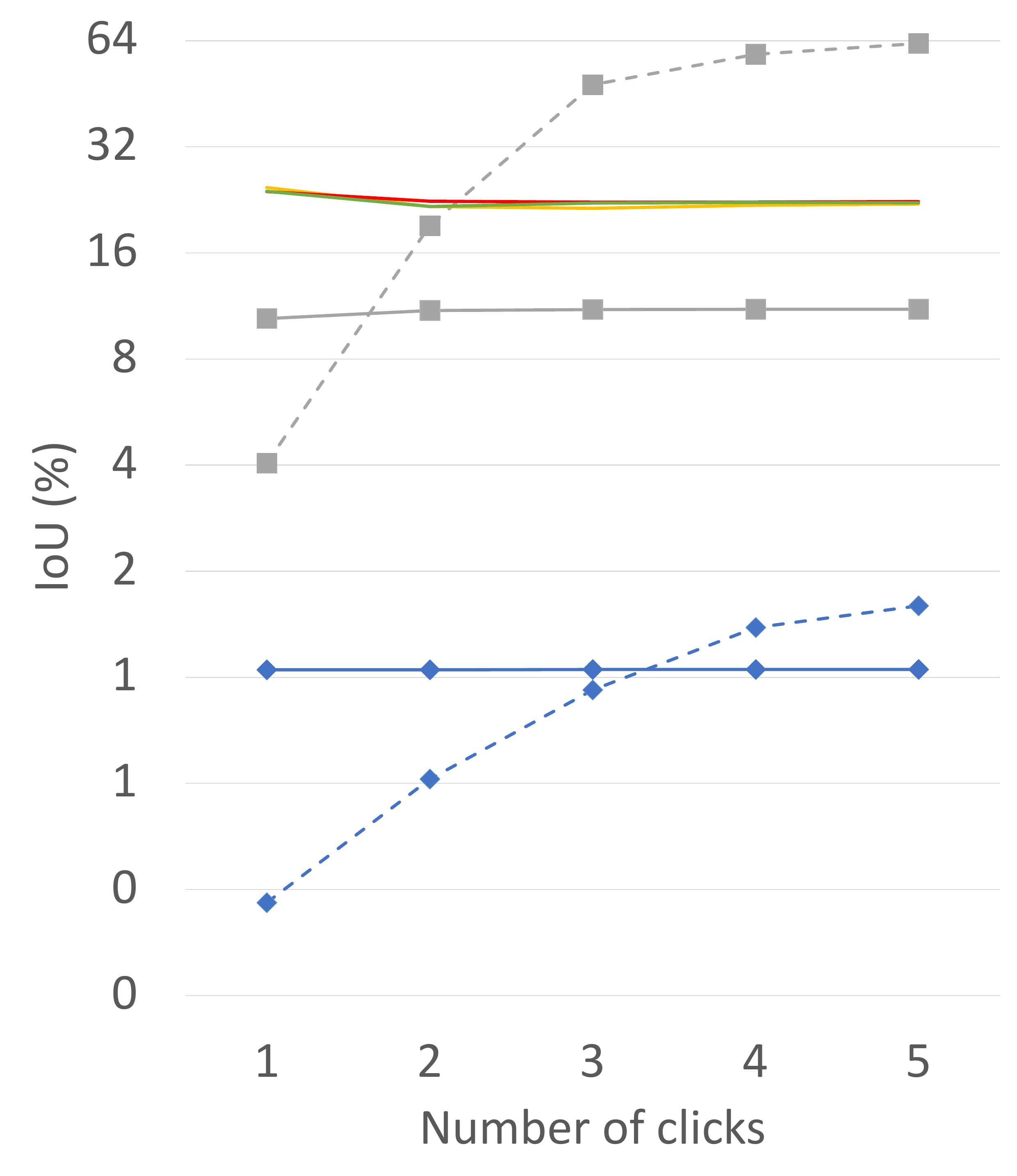}
        \caption{Small (DistractorReal-Val)}
    \end{subfigure}
    \begin{subfigure}[b]{0.49\columnwidth}
        \includegraphics[width=\textwidth]{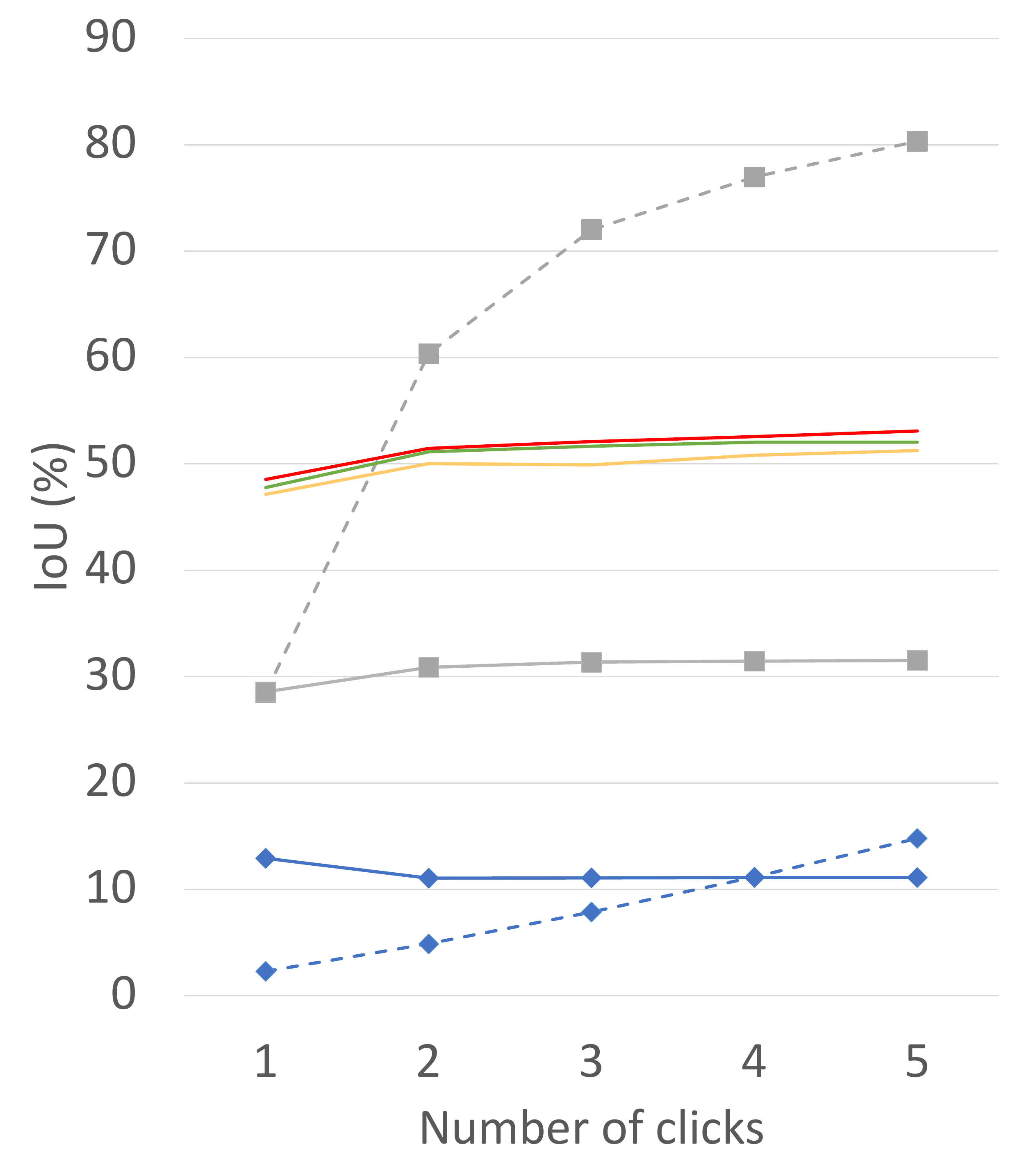}
        \caption{Medium (DistractorReal-Val)}
    \end{subfigure}
    
    \caption{Performance of Interactive Segmentation with different click procedures. Without negative clicks, the current Interactive Segmentation models cannot achieve higher performance when increasing the number of clicks. Our models outperform the model with only one positive click, which better suits the distractor selection task.}
    \label{fig:is_click_sample}
\end{figure}

The \Fref{fig:is_click_sample} shows the performance of Interactive Segmentation models in two different clicking strategies. The public weights are used in the positive-negative strategy, while our finetuned models are tested with positive clicks only. All frameworks using the positive-click strategy, including ours, do not receive large improvements without negative clicks to refine the boundary. In contrast, the performance of RiTM and FocalClick using the positive-negative strategy increases by adding more negative clicks. It demonstrates the existence of negative clicks indeed helps to improve the overall masking performance with multiple clicks from the users.

However, for both RiTM and FocalClick, the existence of negative clicks also does harm to the performance when there is only one single positive click. As shown in \Fref{fig:is_click_sample}, in the first positive click, the new finetuned models achieve higher results than the ones using negative clicks in training. As we mentioned in the main paper, distractors are mostly medium and small objects, and users prefer to use fewer clicks for them. Therefore, the positive-click strategy is more suitable for distractor removal and photo-cleaning applications. Following this core idea and under fair comparison, our framework achieves reasonable performance with one positive click than the other two interactive segmentation models.

\begin{figure*}
    \centering
    \includegraphics[width=\textwidth]{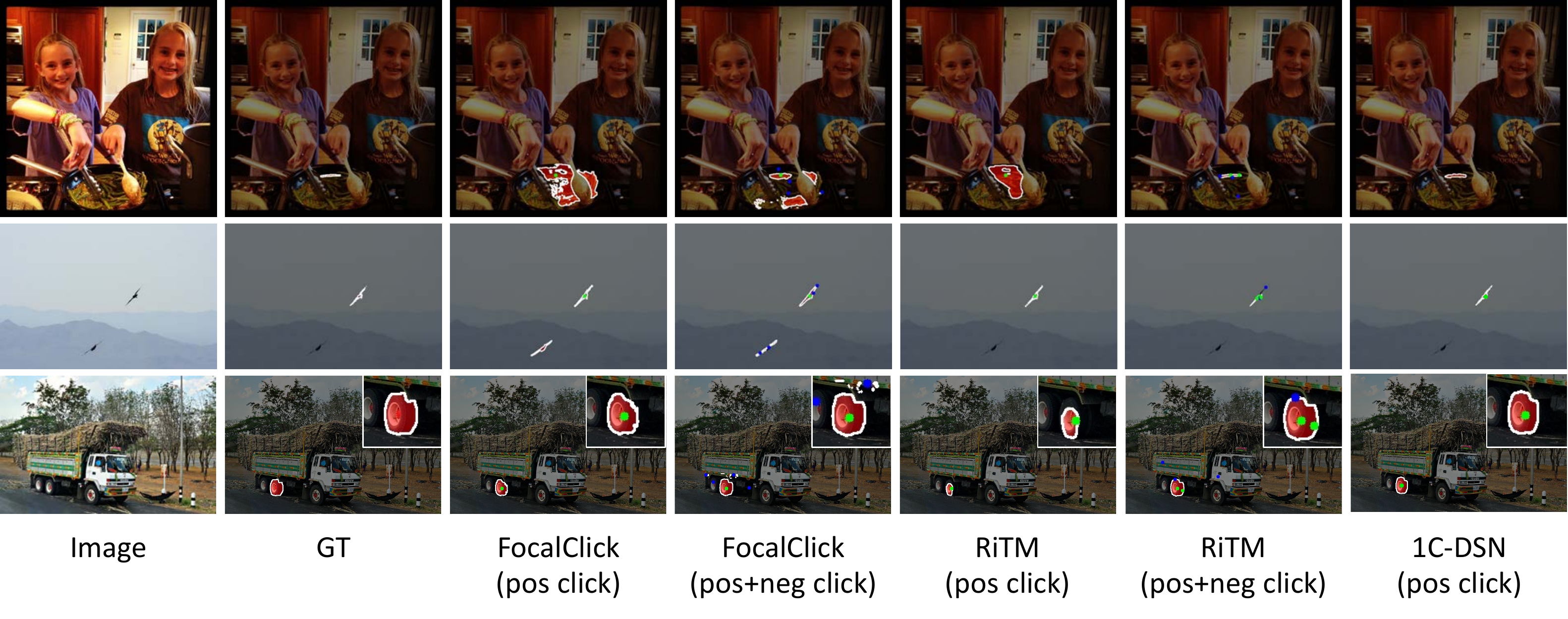}
    \caption{Qualitative results of different click samplers and frameworks on LVIS val set. With only one positive click, our model 1C-DSN can select perfect masks of objects while other frameworks require negative clicks. Except for RiTM using HRNet32, other frameworks use MiT-B3 as backbones. (Green: positive click, Blue: negative click. Best view in digital color)}
    \label{fig:click_compare}
\end{figure*}

Some qualitative results of different click samplers on LVIS val set are shown in \Fref{fig:click_compare}. The backbone used is MiT-B3. Other models with more negative clicks help to improve the detailed segmentation boundary and achieve overall better results. However, our model obtains better masking quality than one-positive-click finetuned FocalClick and RiTM, and requires less user effort to select. The one-click system also helps with group selection scenarios later in the pipeline.

\begin{table}[b]
    \centering
    \small
    \begin{tabu} to \columnwidth {X[2.0, c]X[1, c]X[1, r]X[1.5, c]X[1.5, c]}
     \toprule
     \small{Click Embedding} & IDS & PVM & AP~(\%) & AR~(\%) \\
     \midrule
      & & & 28.9 & 39.2 \\
     & & \checkmark & 29.9 & 39.0 \\
     & \checkmark &  & 23.0 & 42.2 \\
     & \checkmark & \checkmark & 26.7 & 42.0 \\
     \midrule
     \checkmark & & & 34.1 & 41.0 \\
     \checkmark & & \checkmark & 33.7 & 39.0 \\
     \checkmark & \checkmark & & 34.4 & 47.0 \\
     \checkmark & \checkmark & \checkmark & 42.4 & 49.7 \\
     \bottomrule
    \end{tabu}
    \caption{Performance of EntitySeg (SwinL) and 1C-DSN (SwinL) on DistractorSyn-Val set. The click embedding helps producing better exemplar masks then improve the performance in finding similar distractors.}
    \label{tab:click_embedding}
\end{table}

\subsection{Randomness of Clicks}
\begin{figure}[t]
    \centering
    \includegraphics[width=\columnwidth]{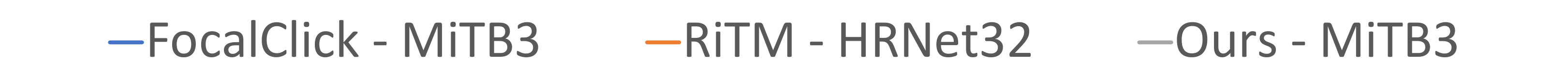}
    \begin{subfigure}[b]{0.49\columnwidth}
        \includegraphics[width=\textwidth]{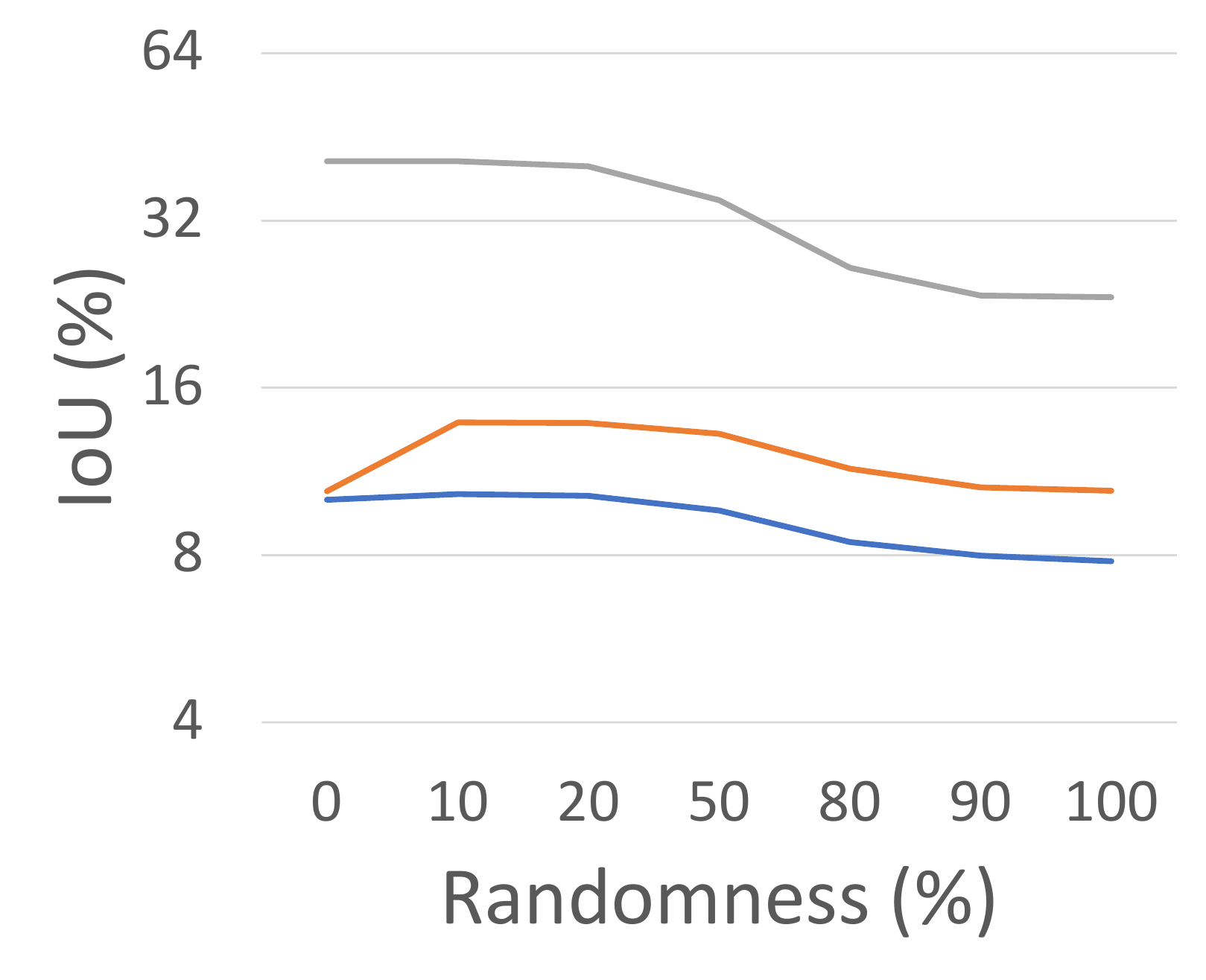}
        \caption{Small (LVIS val)}
    \end{subfigure}
    \begin{subfigure}[b]{0.49\columnwidth}
        \includegraphics[width=\textwidth]{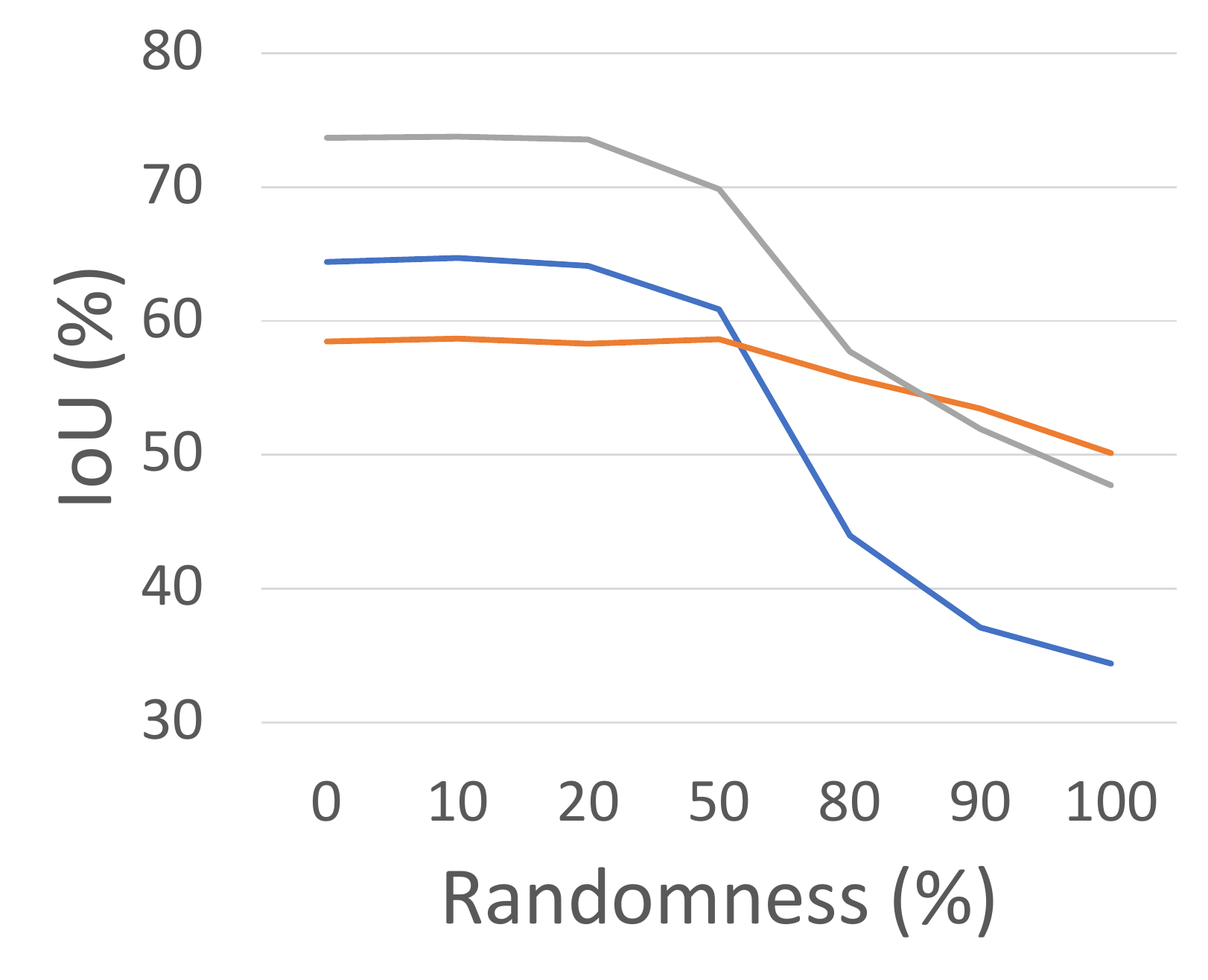}
        \caption{Medium (LVIS val)}
    \end{subfigure}
    
    \caption{The performance of one-positive-click models drops when increasing the randomness of click locations.}
    \label{fig:is_click_random}
\end{figure}

To evaluate the robustness of models with different click positions, we increase the randomness of clicks surrounding the object's center. Let $d_{max}$ be the peak value in the distance map $\Delta$, which localizes in the center of the object. The randomness level $r$ defines a threshold such that the clicks are placed among all positions with $d_i \ge (1.0 - r) \times d_{max}$. When clicks are always at the object center, the randomness $r$ is zero. Otherwise, when the click can be anywhere in the mask, the randomness $r=1.0$. The \Fref{fig:is_click_random} presents the decrease in performance when increasing the randomness of click positions. All models are finetuned on LVIS dataset with one positive click procedure. With small objects, the randomness level does not affect the IoU significantly. However, the performance goes down quickly with medium objects when increasing the randomness from 50\% to 80\%. Our performance still remains high with a large randomness level, indicating the model's robustness to click randomness.

\begin{figure}[t]
    \centering
    \includegraphics[width=\columnwidth]{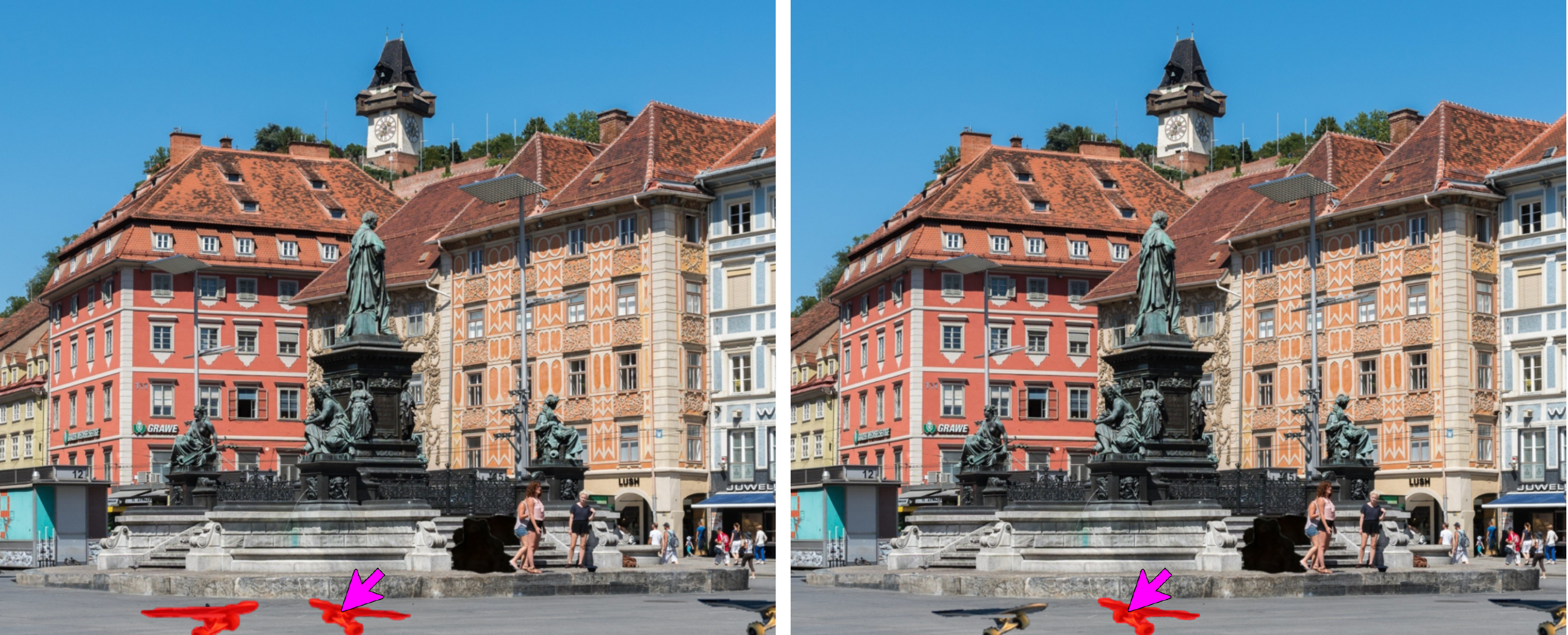}
    \includegraphics[width=\columnwidth]{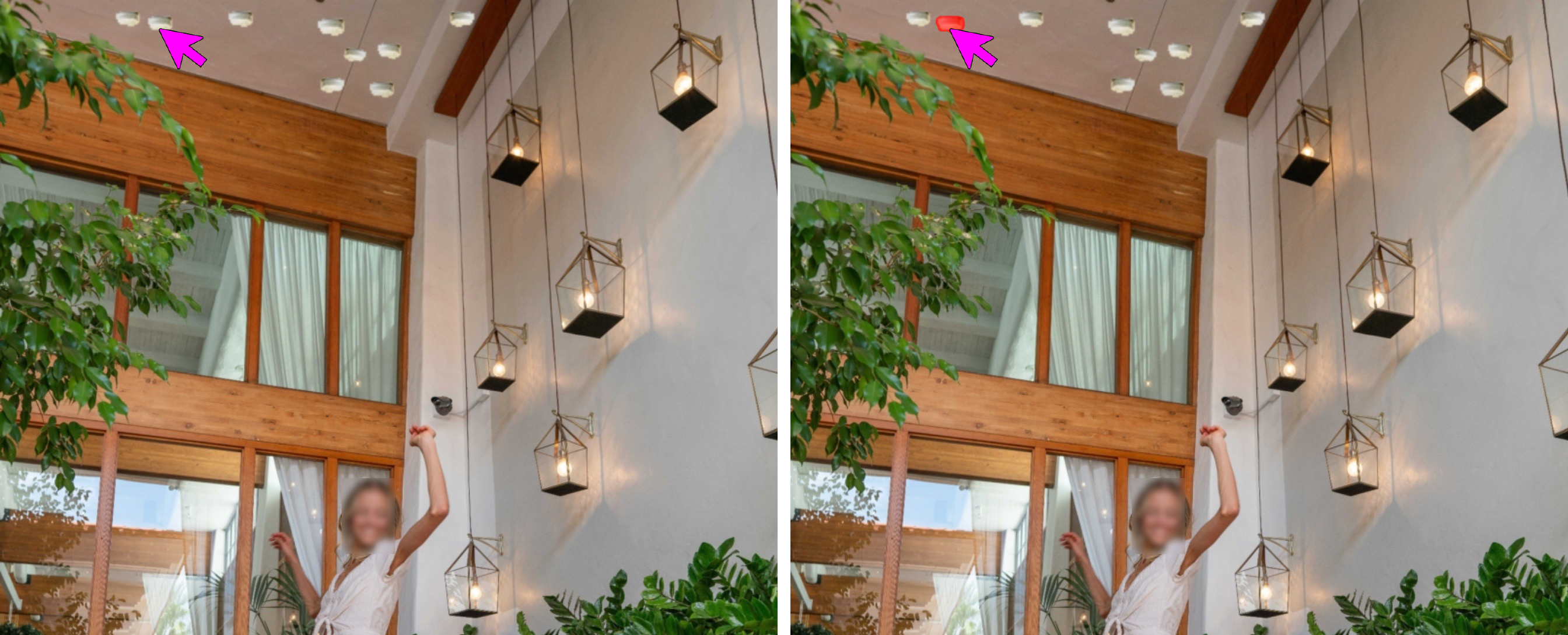}
    \caption{Typical failure cases of EntitySeg (left) without click embedding on DistractorSyn-Val set. EntitySeg baseline may over-predict or under-predict the masks making the self-similarity mining not reliable enough. Our framework (right) still has good predicted mask. Both models using Swin-Large backbone and are trained on Distractor20K. (Best view in color)}
    \label{fig:entseg_failure}
\end{figure}

\subsection{Similarity Findings without Click Embedding}

We show in \Tref{tab:click_embedding} the performance of EntitySeg~\cite{qi2021open} model in similarity finding and group selection. This experiment aims to ensure that our one-click-based segmentation model is necessary for the group selection scheme. 

We simply apply our CPN and PVM modules to an EntitySeg model (without click embedding inputs) finetuned on Distractor20K dataset. We use the model predictions on one image to extract exemplar masks for the CPN and PVM modules. For the clicks which do not have corresponding masks, no further steps are executed and then the prediction output will be empty. Without running IDS or PVM, the EntitySeg model provides lower recall and precision than our models. The \Fref{fig:entseg_failure} gives typical failure cases of EntitySeg models. Masks can be over-segmented or wrongly detected by EntitySeg without click inputs. Therefore, adding either IDS or PVM steps does not improve the final results for EntitySeg baseline model.


\section{Compare CPN with Other Self-similarity Methods}

\begin{figure*}[h!]
    \centering
    \includegraphics[width=\textwidth]{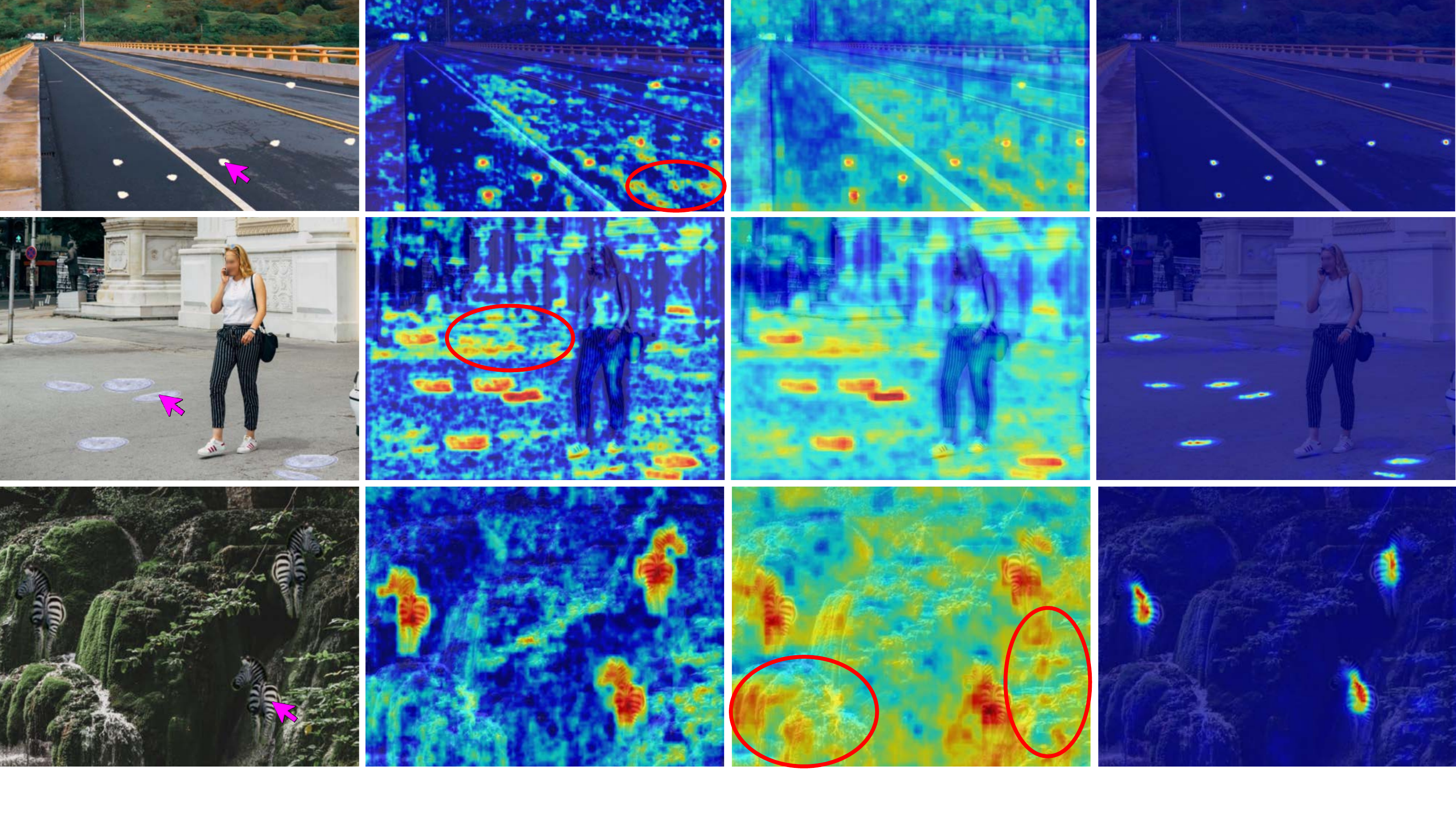}
    \includegraphics[width=\textwidth]{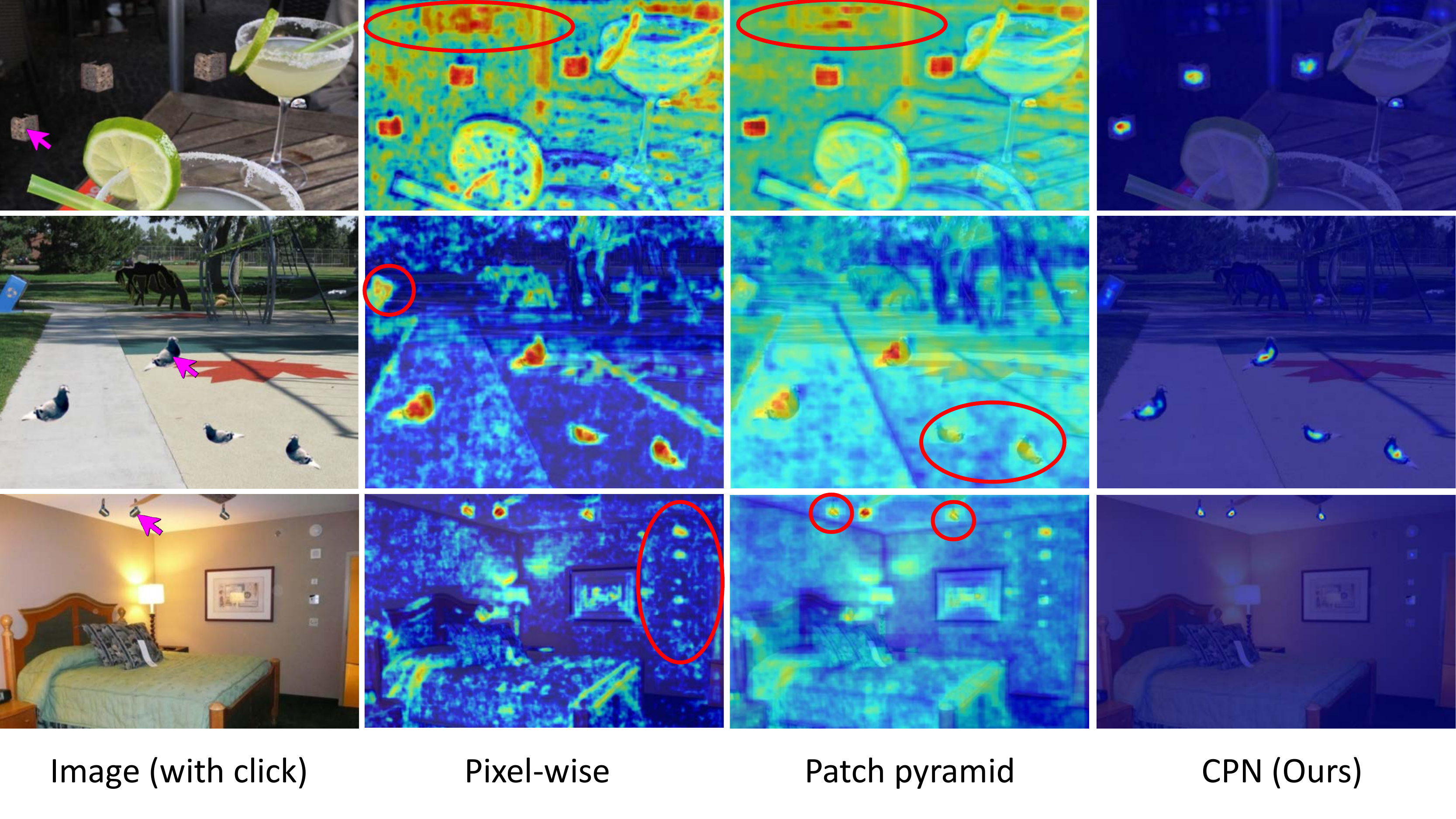}
    
    \caption{Our framework produces cleaner heatmaps than other naive self-similarity methods. Red eclipses indicate the false or missed detection regions. Images are from DistractorSyn-Val dataset. (Best view in color)}
    \label{fig:hm_compare}
\end{figure*}

\paragraph{Related Works.} Self-similarity is a commonly used technique in vision tasks to find repeating patterns and learn better globally consistent features. It has been used in various models, including non-local network \cite{nonlocal}, contextual attention \cite{contextual}, self-attention \cite{selfattention, attentionneed}, and transformer-related models \cite{vit, swin, pvt, segformer}. Attention has also shown to benefit almost all vision models, including image super-resolution \cite{srattention}, object detection \cite{deformabledetr,Wang_2022_CVPR}, and image synthesis \cite{esser2021taming, Dinh_2022_CVPR, Ho_2021_ICCV}, among others~\cite{Pham_2021_CVPR,Saini_2022_CVPR,pham2022improving, saini2022recognizing, Tran_2021_CVPR}. The most similar work related to our task is visual counting \cite{count1,count2, count3, count4, Ranjan_2018_ECCV}, which aims to localize all similar objects within the same images by actively sampling a few of them. However, visual counting works do not require masking the objects, and the targets of visual counting are usually the main subjects of the photos. In our task, we may face more complicated and challenging image contexts, where diverse context yields a high false positive detection rate. To address this issue, we leverage a transformer decoder to learn cross-scale attention and generate the attention heatmap, along with an additional verification scheme to remove false positives.

\paragraph{Point Detection Precision and Recall.} To evaluate the performance of the CPN module in similarity heatmap generation, we use Area Under the Curve of Precision-Recall (AUC-PR) on the similarity heatmap. The metric is used in Table {\color{red}4}. A click located at the ground-truth mask region is counted as a true positive; otherwise, it is counted as a false positive. The precision is the proportion of true positive clicks and the total predicted clicks. The recall is the ratio between the number of masks having predicted clicks over the total of masks. We compute precision and recall in different thresholds to get the curve between them. 

\begin{figure}[t!]
    \centering
    \includegraphics[width=\columnwidth]{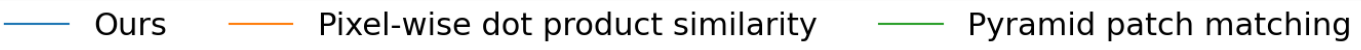}
    \includegraphics[width=0.7\columnwidth]{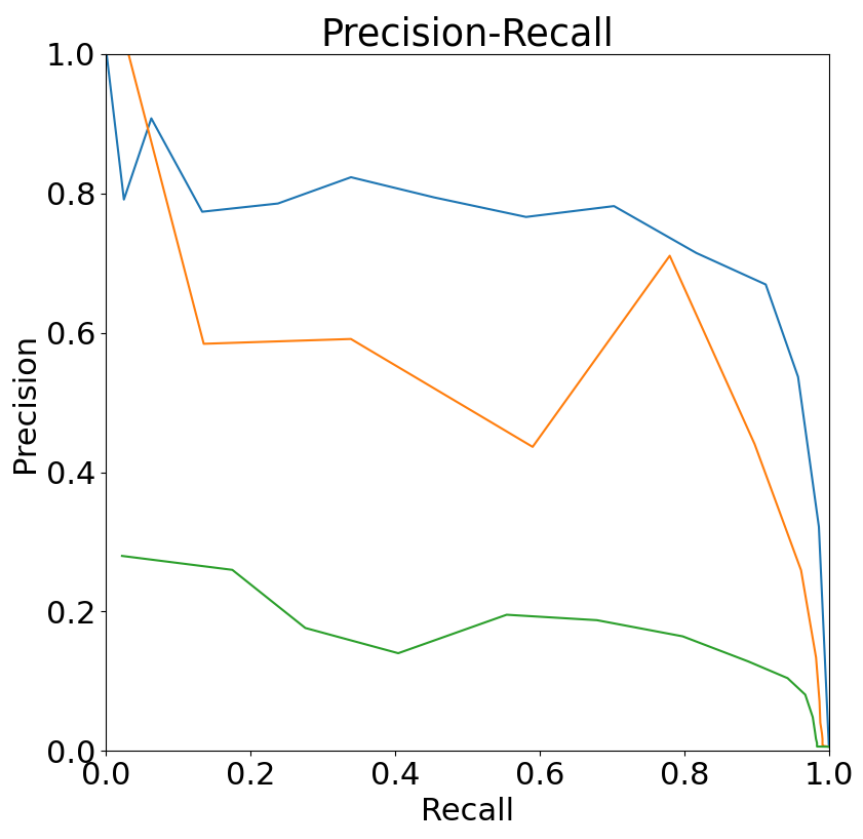}
    
    \caption{Precision-Recall of heatmaps predicted by different similarity finding approaches on DistractorSyn-Val. Our models has higher precision than other methods.}
    \label{fig:similarity_compare}
    \vspace{-1em}
\end{figure}

\begin{figure}[b!]
    \centering
    \begin{subfigure}[b]{0.49\columnwidth}
        \includegraphics[width=\textwidth]{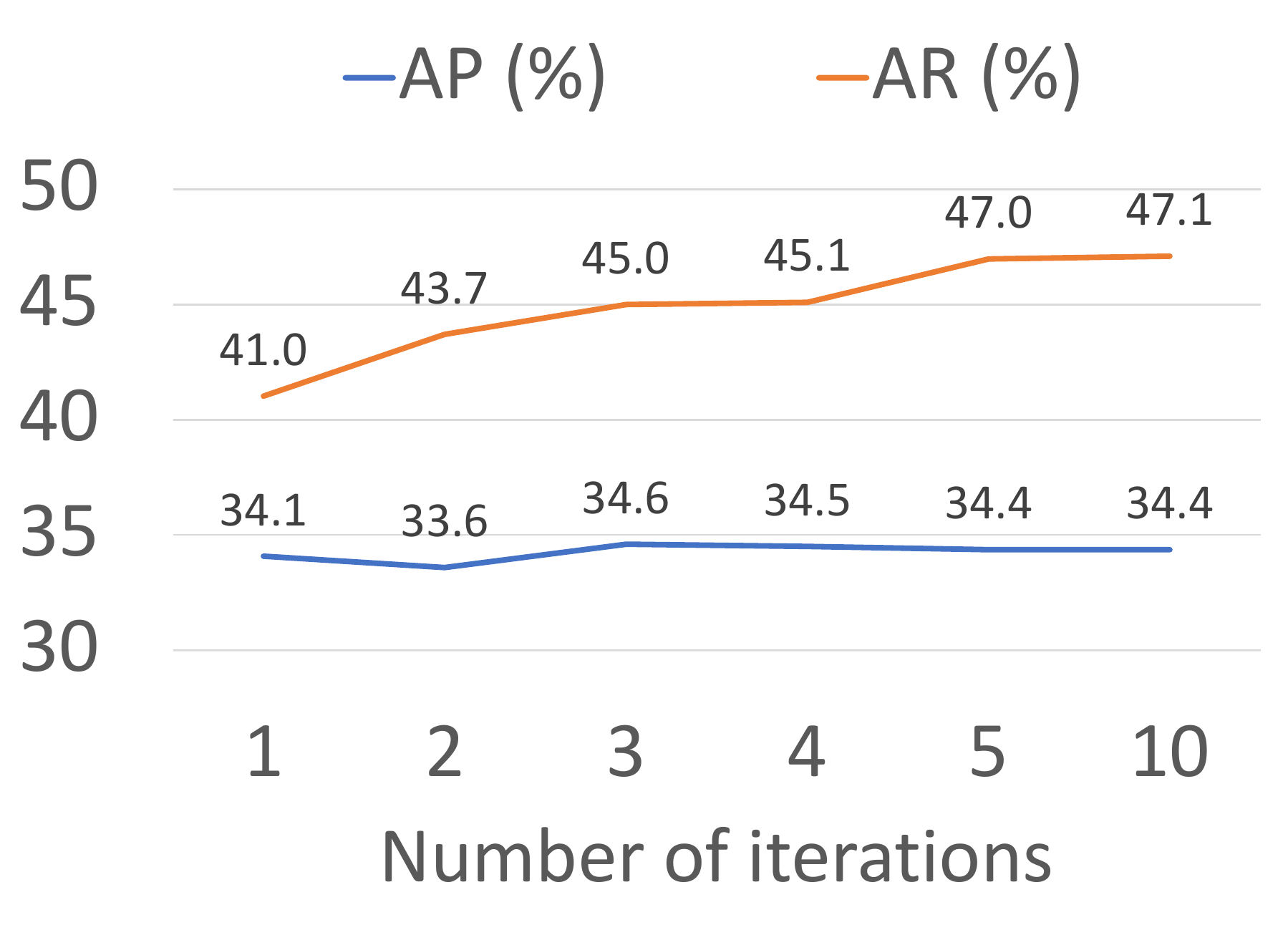}
        \caption{Average Precision Recall}
    \end{subfigure}
    \begin{subfigure}[b]{0.49\columnwidth}
        \includegraphics[width=\textwidth]{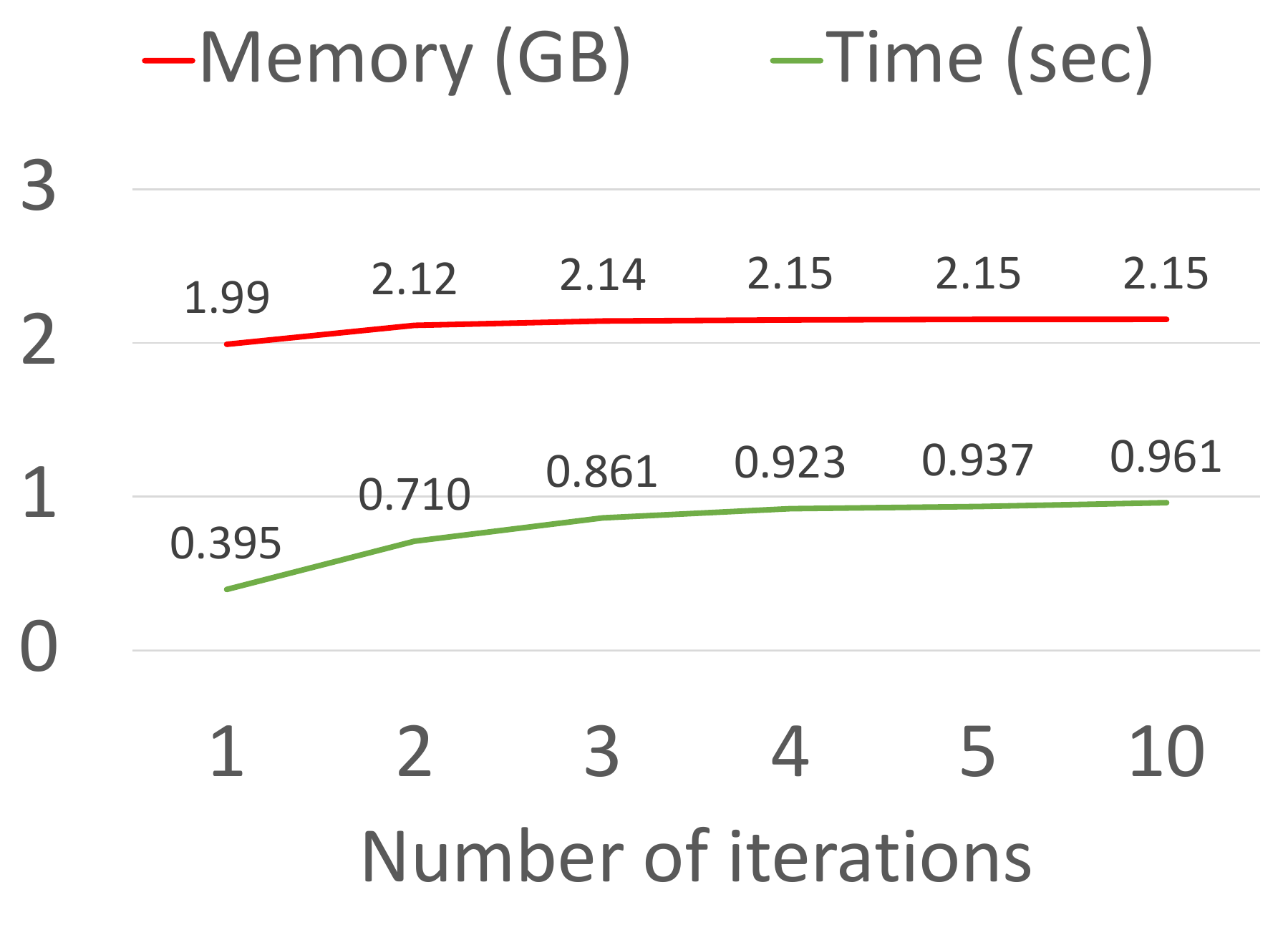}
        \caption{Computational cost}
    \end{subfigure}
    \caption{The increasing number of iterations improves the Average Recall while maintaining the Average Precision. The computational cost is also increased. The performance is saturated after about five iterations.}
    \vspace{-1em}
    \label{fig:ids_iters}
\end{figure}

\paragraph{Pixel-wise Dot Product Similarity.} Since there are no previous works on distractor similarity findings and it is not fair enough to directly compare with visual counting works, we can only compare our CPN with some naive baselines. With the finer feature map $X_1 \in \R^{h \x w \x d}$ and the mask $M$, we compute the query vector $q \in \R^d$ by Masked Global Average Pooling~\cite{zhang2020sg}. The dot product similarity is then computed between $X_1$ and $q$ to get the heatmap $H \in \R^{h \x w}$.


\paragraph{Pyramid Patch Matching.} We firstly build the 3-level pyramid features of $X_1$ with the scales $\frac{1}{4}, \frac{1}{8}, $ and $\frac{1}{16}$ of the original image. With the query mask, the query patch feature $q\in \R^{3\x 3 \x d}$ is extracted by RoI-Align. By sliding the query patch feature on the feature pyramid, we can compute the similarity at each location to the query patch with Sum Squared Distance (SSD). The final heatmap is the average of responses of all pyramid levels.

\paragraph{Comparing with our CPN.} \Fref{fig:hm_compare} shows the differences in heatmaps produced by different methods. Pixel-wise similarity can cause many false positives, while the patch pyramid approach is not robust to objects with variant appearances. Our proposed method generates cleaner heatmaps with high precision. The precision-recall curve of three methods on DistractorSyn-Val is shown in \Fref{fig:similarity_compare}. Our method outperforms other baselines with the balance between precision and recall rate.

\begin{figure}[t!]
    \centering
    \begin{subfigure}[b]{0.49\columnwidth}
        \includegraphics[width=\textwidth]{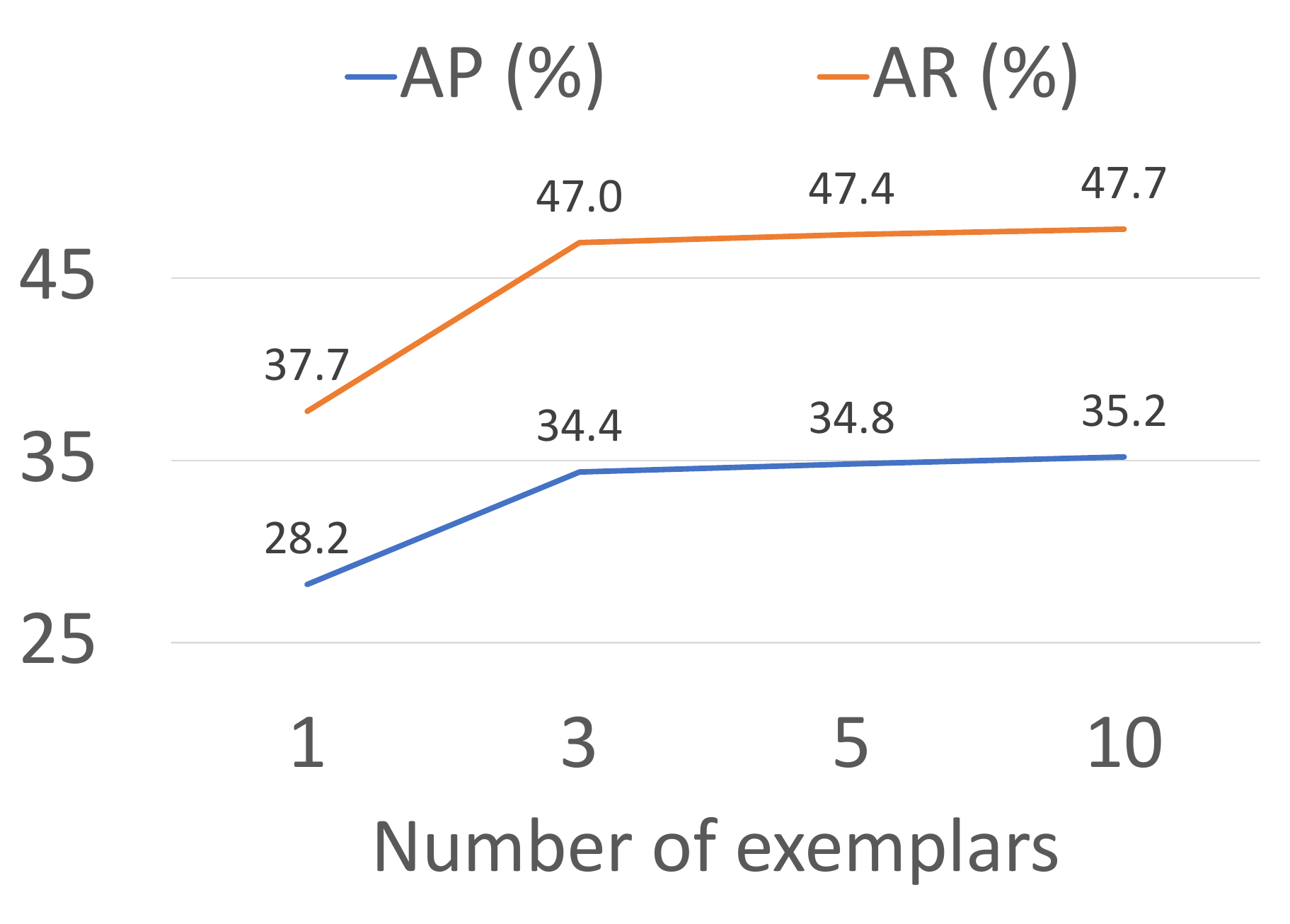}
        \caption{Average Precision Recall}
    \end{subfigure}
    \begin{subfigure}[b]{0.49\columnwidth}
        \includegraphics[width=\textwidth]{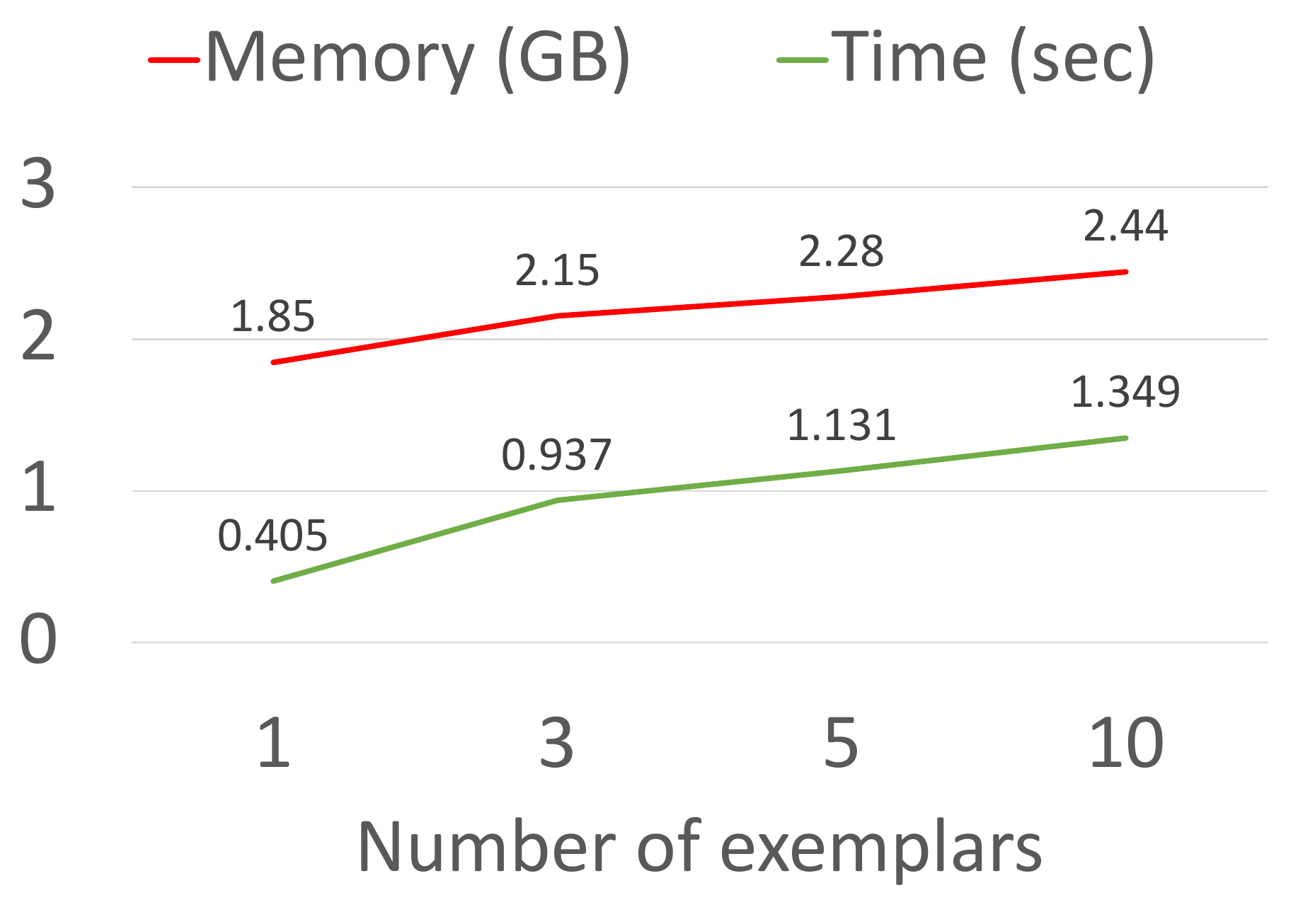}
        \caption{Computational cost}
    \end{subfigure}
    \caption{The performance of IDS increases proportionally with the number of exemplars and the computational cost.}
    \label{fig:ids_exemplars}
\end{figure}

\begin{figure}[b]
    \centering
    \begin{subfigure}[b]{0.49\columnwidth}
        \includegraphics[width=\textwidth]{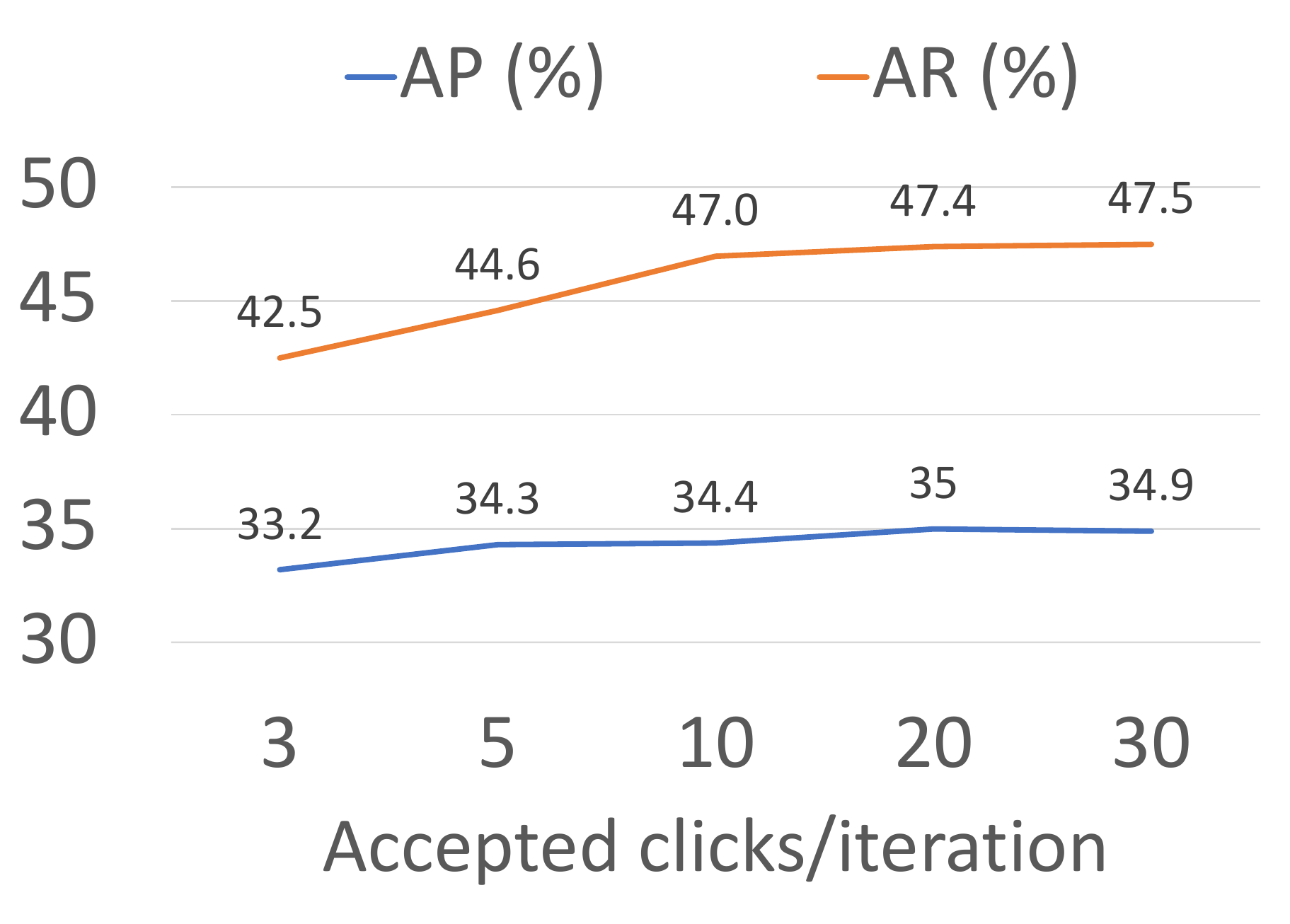}
        \caption{Average Precision Recall}
    \end{subfigure}
    \begin{subfigure}[b]{0.49\columnwidth}
        \includegraphics[width=\textwidth]{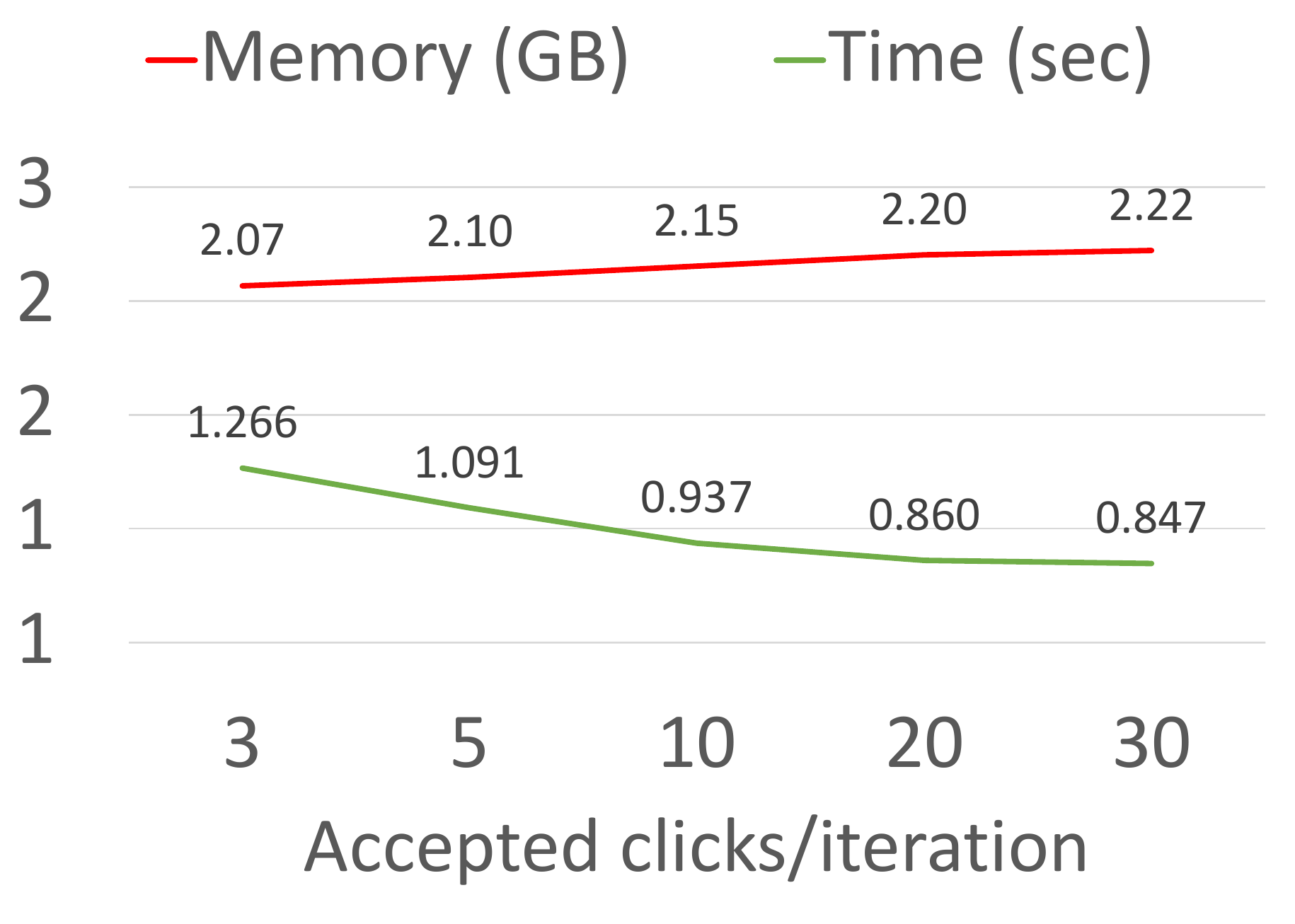}
        \caption{Computational cost}
    \end{subfigure}
    \caption{Accepting more clicks in each iteration improves the performance and the speed. However, memory usage also grows as a result.}
    \label{fig:ids_topk}
\end{figure}

\begin{figure*}[t!]
    \centering
    \includegraphics[width=\textwidth]{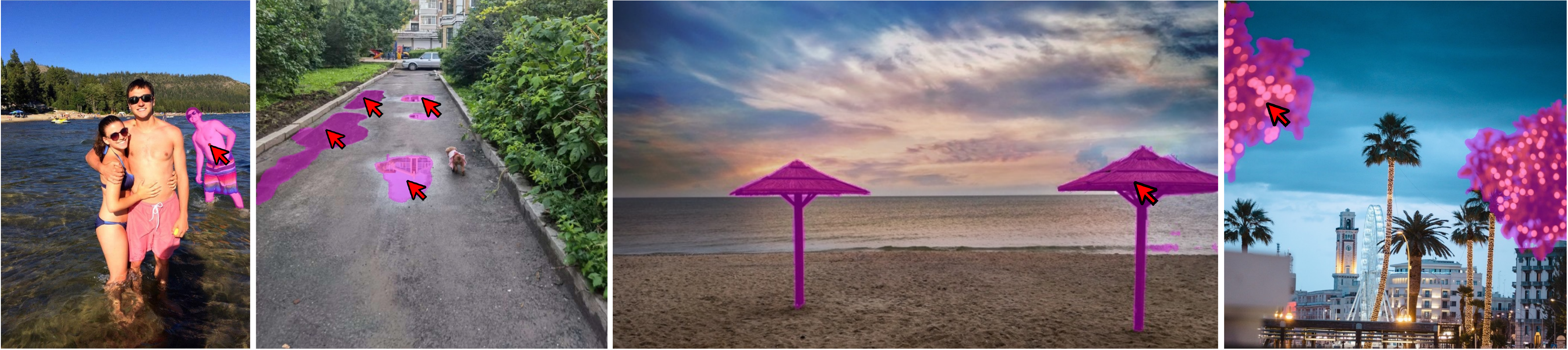}
    \caption{Our model also performs well with large objects.}
    \label{fig:large}
\end{figure*}

\begin{figure*}[t!]
    \centering
    \includegraphics[width=\textwidth]
    {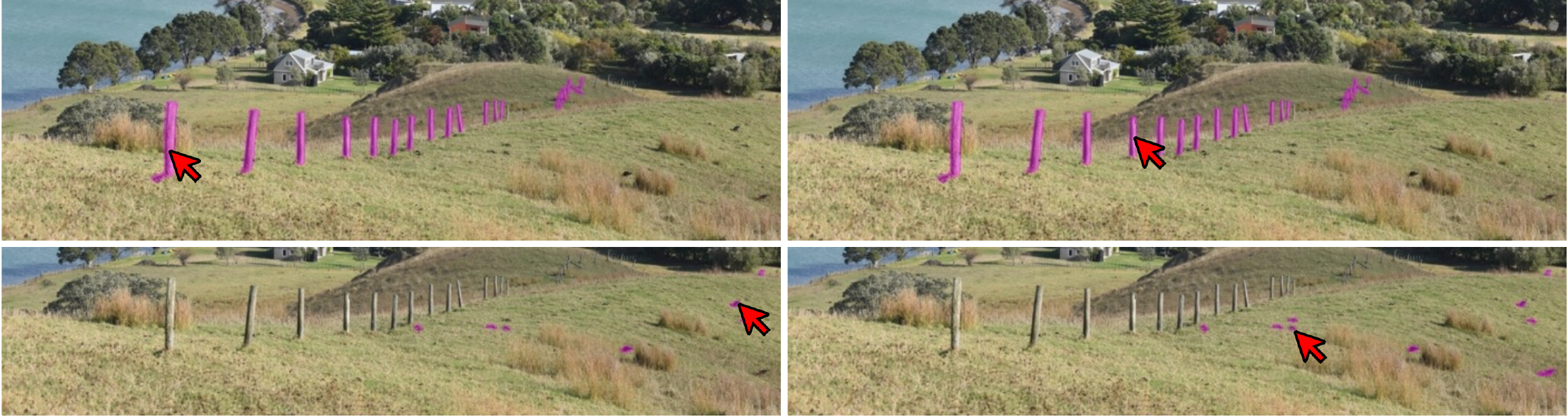}
    \caption{The robustness of our model when clicking on different objects.}
    \label{fig:consistency}
\end{figure*}

\section{Hyper-parameters of IDS}
The following sections evaluate the performance of our proposed IDS with different hyper-parameters on DistractorSyn-Val. All experiments are with the SwinL backbone and trained on DistractorSyn14K. The pretrained weights on DistractorReal20K are used for feature extraction and mask generation modules. If not explicitly stated, all experiments have default hyper-parameters: the number of iterations $N=5$, the number of exemplars $m=3$, and the number of accepted clicks for each iteration $k=10$. To make the comparison consistent, the PVM does not validate the outputs at each iteration.

Besides the Average Precision (AP) and Average Recall (AR), we also compute the time and GPU memory complexity of different hyper-parameters. While the time is measured before starting IDS until the last iteration, the GPU memory is the additional cost raised by the IDS, not by the whole network. The memory amount is computed with PyTorch API.

\subsection{Number of Iterations}

The \Fref{fig:ids_iters} illustrates the performance of IDS with different numbers of iterations $N$. When $N=1$, all proposal clicks are accepted, which are equivalent to the non-IDS experiment. Other experiments use the default value of $k=10$ by default. There is a trade-off between computational cost and the performance of the framework when increasing the number of iterations. The AR increases proportionally to $N$ until the fifth iteration. Because almost all clicks have been accepted after five iterations, continuously running the CPN after that only yields incremental improvements. 

\subsection{Number of Accepted Clicks}

The results of different accepted clicks for each iteration are shown in \Fref{fig:ids_topk}. The performance and speed of the entire IDS process increase When accepting more clicks for each round. However, it also consumes more memory for mask generation. In practice, depending on the occurrence of distractors in the image, we can balance between the number of accepted clicks and the number of iterations to achieve the best results.

\subsection{Number of Exemplars}
We change the number of exemplars used for querying similar objects in each iteration of IDS process. The results are shown in \Fref{fig:ids_exemplars}. An increasing in the number of exemplars significantly rises the time and memory complexity. Additionally, the AP and AR are also improved with more exemplars.

\section{Additional Qualitative Results}

\begin{figure*}[t!]
    \centering
    \includegraphics[width=\textwidth]{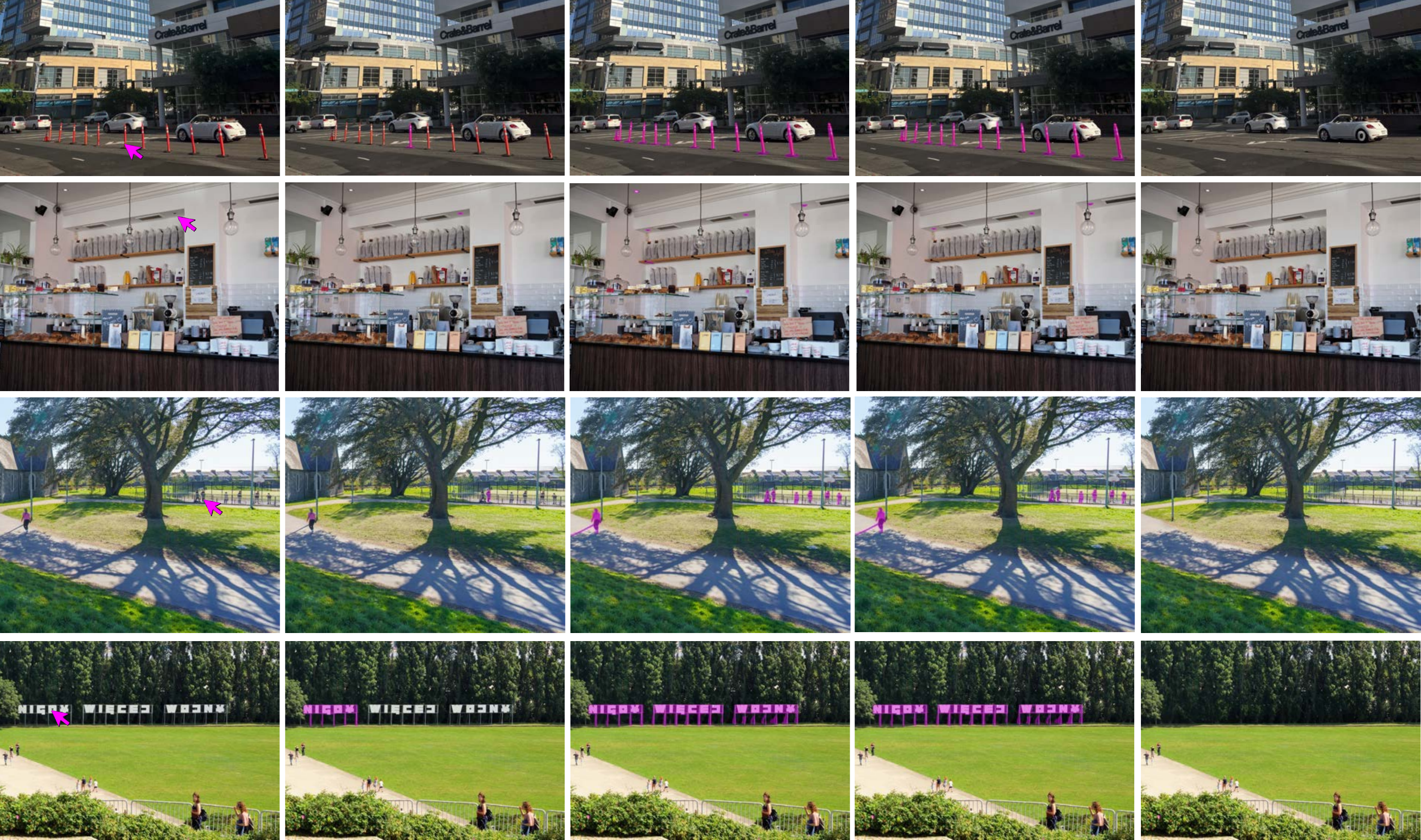}
    \includegraphics[width=\textwidth]{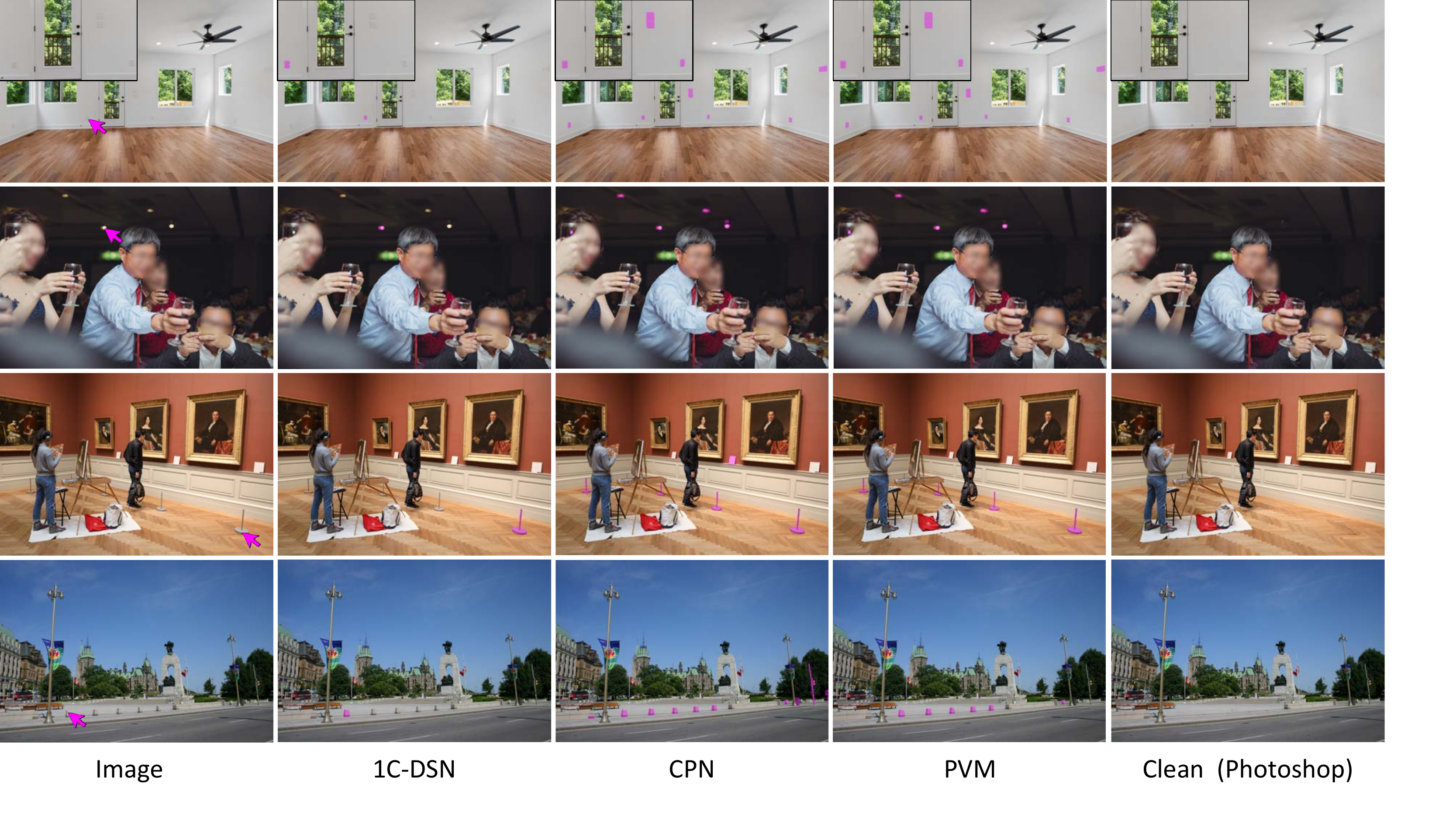}
    \caption{Some intermediate results without IDS process. Our CPN successfully finds similar objects with a high recall rate, and PVM correctly removes false positives to clean. (Best view in color and zoom-in)}
    \label{fig:results_noiter}
\end{figure*}



\begin{figure*}[t!]
    \centering
    \includegraphics[width=\textwidth]{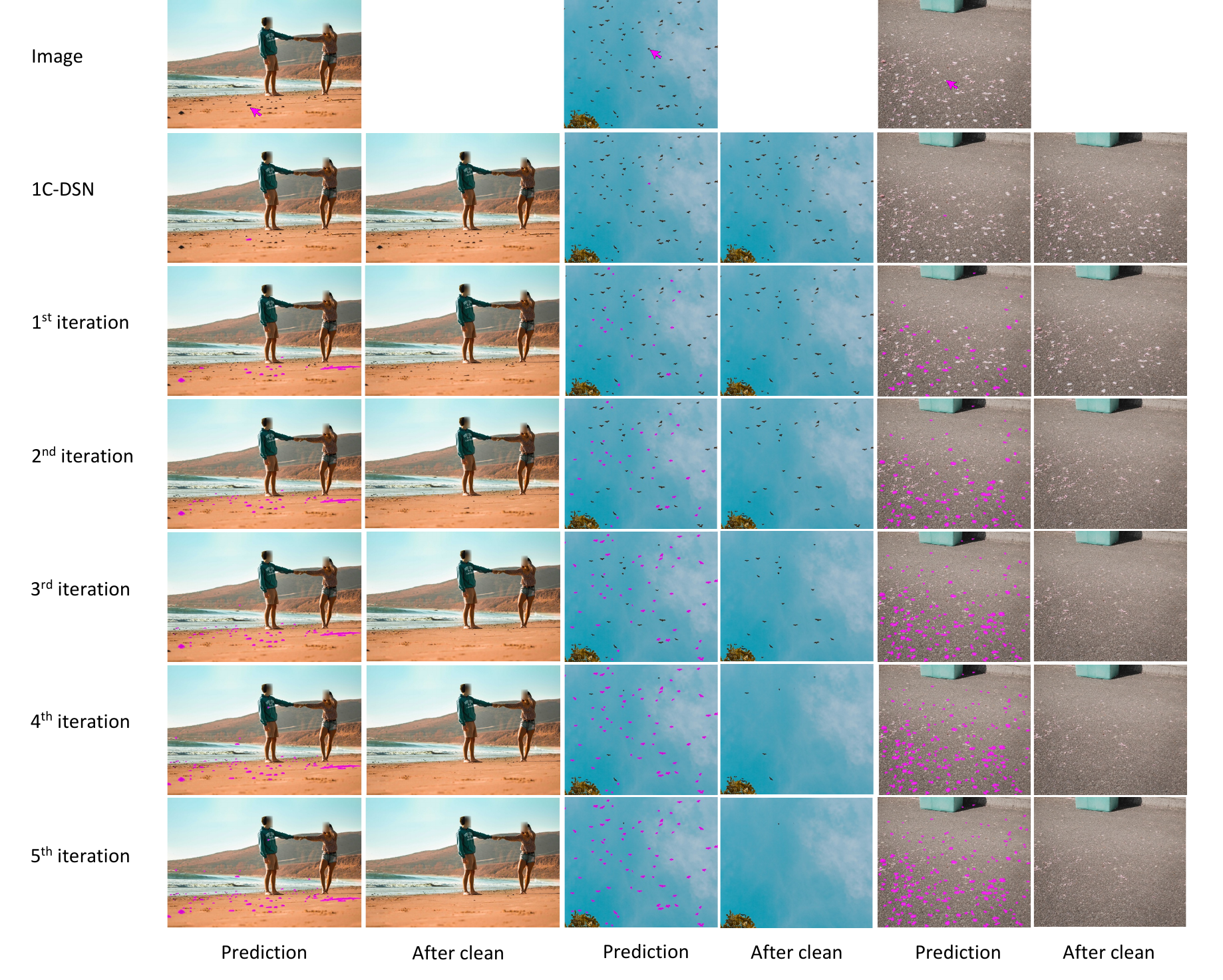}
    \caption{The progress of cleaning photos with IDS. For each iteration, some new similar distractors are detected, and the photos become cleaner than in the previous step. In the end, all distractors are removed from the image. (Best view in color and zoom-in)}
    \label{fig:results_ids}
\end{figure*}

Our framework not only works with tiny distractors but also yield good results on large objects. \Fref{fig:large} shows some examples where the selecting objects are larger than $10\%$ of the image. Besides, \Fref{fig:consistency} illustrates the robustness of our model in which the results are consistent while different objects are clicked. 

Some additional results on real photos are shown in the \Fref{fig:results_noiter} and \Fref{fig:results_ids}. In some simple cases where there are not many repeated distractors, the CPN and PVM frameworks can work perfectly without IDS process. The CPN tries to achieve a high recall rate, and then the PVM helps increase precision by removing outliers. 

The \Fref{fig:results_ids} shows some extreme cases where many similar distractors appeared, and the IDS joins in selecting all similar distractors with only one click. The photos get cleaner after each iteration because more distractors are selected and removed.


{\small
\bibliographystyle{ieee_fullname}
\bibliography{supplementary}
}

%% file: sections/0_abstract.tex
\begin{abstract}
\vspace{-2.5em}
In photo editing, it is common practice to remove visual distractions to improve the overall image quality and highlight the primary subject. However, manually selecting and removing these small and dense distracting regions can be a laborious and time-consuming task. In this paper, we propose an interactive distractor selection method that is optimized to achieve the task with just a single click. Our method surpasses the precision and recall achieved by the traditional method of running panoptic segmentation and then selecting the segments containing the clicks. We also showcase how a transformer-based module can be used to identify more distracting regions similar to the user's click position. Our experiments demonstrate that the model can effectively and accurately segment unknown distracting objects interactively and in groups. By significantly simplifying the photo cleaning and retouching process, our proposed model provides inspiration for exploring rare object segmentation and group selection with a single click. More information can be found at \url{https://github.com/hmchuong/SimpSON}.
\end{abstract}
\vspace{-1em}

%% file: sections/1_introduction.tex
\section{Introduction}
\footnotetext[1]{This work was done during Chuong Huynh's internship at Adobe}
Both professional photographers and casual users often require efficient photo retouching to enhance the quality of their images. One essential aspect of this task is the removal of visual distractions from photos \cite{fried2015finding}. These distractions can take various forms, such as unexpected pedestrians, objects that are cropped out of the photo's edge, dirty spots on the ground, repeated outlets on a wall, or even colorful and blurry lens flare. These distractions can be challenging to categorize due to their diverse appearance. As a result, users tend to select and mask them entirely and use photo editing software such as Photoshop to remove them.

Segmentation is necessary for photo cleaning tasks because rough masks may not be suitable for all scenarios. Accurate masks are required in situations where distractors are touching the main foreground subjects or where distractors are small but dense in the image. User-drawn rough masks can result in the deletion of too much background texture when connected. In other cases, users may have a mask that covers the entire object but does not change the background too much. In all scenarios, our findings suggest that for inpainting, a tiny dilation from a highly accurate mask produces better background preservation and fewer leftover pixels of distractors. This finding is consistent with most of the existing inpainting models.

The process of manually masking distracting elements in a photo can be a tedious and time-consuming task. Users often seek an automated tool that can efficiently select and segment all distractors. One approach is to train an instance segmentation model like Mask-RCNN \cite{he2017mask} to detect and segment distractors in a supervised manner. However, identifying distractors can be subjective, and collecting datasets requires scientific validation of the distractor annotations to ensure that most users agree. For instance, Fried \textit{et al.} \cite{fried2015finding} invited 35 users to mark distractors on a single image and received varying feedback. Even with a model that detects distractors, it may not always satisfy users' preferences. Therefore, tasks like these should rely heavily on user interaction, such as allowing users to click and decide where to retouch photos based on their own preferences.

Our goal is to propose a single-click distractor segmentation model. With the rapid development of panoptic segmentation technologies like PanopticFCN \cite{li2021fully} and Mask2Former \cite{cheng2022masked}, can we utilize state-of-the-art models to retrieve distractor masks by clicking on the panoptic segmentation results? Unfortunately, most distractors belong to unknown categories, and some are tiny, making them difficult to segment using models~\cite{huynh2021progressive, bui2022multi} trained on datasets such as COCO \cite{lin2014microsoft}, ADE20K \cite{zhou2017scene}, or Cityscapes \cite{cordts2015cityscapes} with a closed-set of categories. Qi \textit{et al.} proposed entity segmentation \cite{qi2021open} to train panoptic segmentation in a class-agnostic manner to address the long-tail problem, but it still may not be guaranteed to separate all regions in the photos.

What if we use clicks as the input guidance for segmentation? Interactive segmentation models are closely related to our task, and recent works like FocalClick \cite{chen2022focalclick} and RiTM \cite{sofiiuk2022reviving} have achieved practical and high-precision segmentation performance. However, interactive segmentation aims to use multiple clicks, including positive and negative ones, to segment larger foreground objects accurately, especially the boundary regions. In our task, we focus more on medium to small distracting objects and only require a single positive click to select semi-precise masks for inpainting purposes. The difference in our goal makes it challenging to follow the problem definition of interactive segmentation. Additionally, previous interactive segmentation models cannot select objects in groups, whereas most of our distractors are repeated, dense, and evenly distributed across photos.

This paper addresses the two challenges of accurate one-click universal class-agnostic segmentation and efficient similarity selection. Our proposed method can significantly reduce the photo retouching process from hours (e.g., 100+ clicks) to minutes (e.g., 1-2 clicks) when removing dense and tiny distractors. Firstly, we optimize the click-based segmentation model to accurately segment distractor-like objects with a single click. This is achieved by utilizing the entity segmentation \cite{qi2021open} method to discard category labels and using single-click embedding to guide the segmentation of a single object. Secondly, we design a transformer-based Click Proposal Network (CPN) that mines similar distractor-like objects within the same image and regress click positions for them. Lastly, we rerun the single-click segmentation module using the proposed clicks to generate the mask and verify the similarity among the selected objects via the Proposal Verification Module (PVM). We also run the process iteratively to ensure that more similar objects are fully selected. In summary, our contributions consist of three main aspects: 

\begin{itemize}
\item We introduce a novel one-click Distractor Segmentation Network (1C-DSN) that utilizes a single-click-based approach to segment medium to small distracting objects with high accuracy. Unlike other interactive segmentation methods, our model targets the segmentation of distracting objects with just one positive click. Our model is capable of generalizing well to objects of any rare categories present in the photos.

\item We propose a Click Proposal Network (CPN) that mines all similar objects to the user's single click. The proposed clicks are then reused in the segmentation model, and their similarity is verified using the Proposal Verification Module (PVM). This allows for the group selection of distracting objects with one click.

\item We further explore running the selection process iteratively to fully select similar distractors with slightly diverse appearances. Our proposed distractor selection pipeline, which we call 'SimpSON,' significantly simplifies the photo retouching process. By using SimpSON, users can remove distracting objects in their photos quickly and easily with just a few clicks.
\end{itemize}

%% file: sections/2_related.tex
\section{Related works}
\paragraph{Visual Distraction in Photography}
Visual distracting elements in photos are elements that attract users' attention but are not the primary subject of the photo. However, according to~\cite{fried2015finding}, the saliency map \cite{c1,c2,d2,d3,d4} may not be highly correlated with visual distractors because the main subject usually has the peak in the attention map. Although efforts have been made to detect and retouch scratches \cite{oldphoto}, noise, and dirty dots in photos, and automatic and interactive face retouching~\cite{faceretouching} has already been widely deployed in commercial products, only a few research works~\cite{aberman2022deep} have targeted automatic general distractor detection and editing due to the high variance of distractor categories and appearances. In this work, our aim is to develop an interactive distractor selection and masking method in photos, along with automatic grouping and selection of all similar distractors.

\paragraph{Interactive Segmentation}\label{sec:iteractive}
Interactive segmentation involves allowing users to provide a small amount of interaction to complete the target segmentation. Xu \textit{et al.} \cite{xu2016deep} proposed the first deep learning-based segmentation and introduced positive and negative clicks as inputs. BRS \cite{brs}, and f-BRS \cite{fbrs} introduced online learning to optimize the segmentation results, while FCA-Net \cite{fcanet} by Lin \textit{et al.} focuses more on the initial click and uses feature attention to improve the segmentation results. RiTM \cite{sofiiuk2022reviving} generates the following segmentation by fully utilizing the masking results from previous iterations, while CDNet \cite{cdnet} presented how to use self-attention to propagate information among positive and negative clicks. FocalClick \cite{chen2022focalclick} revisited a series of interactive segmentation techniques and proposed to use local inference for a more efficient and deployment-friendly network. In this paper, we draw from the experience of interactive segmentation to use clicks as user inputs. However, due to the nature of distractor removal tasks in photo retouching and cleaning use cases, users prefer to use an undo operation if the model over-predicts the mask, instead of switching between positive and negative clicks. Additionally, distractors are usually smaller than foreground objects, so we redefined our task with only positive clicks and optimized the model with fewer positive clicks. Furthermore, previous works did not allow users to make group selections via self-similarity mining, while it is a highly demanded user need for distractor removal, which we address in our proposed method.


%% file: sections/3_methodology.tex
\section{Methodology: SimpSON}

\begin{figure*}[t]
\centering
\vspace{-1em}
\captionsetup{type=figure}
\includegraphics[width=1.\linewidth]{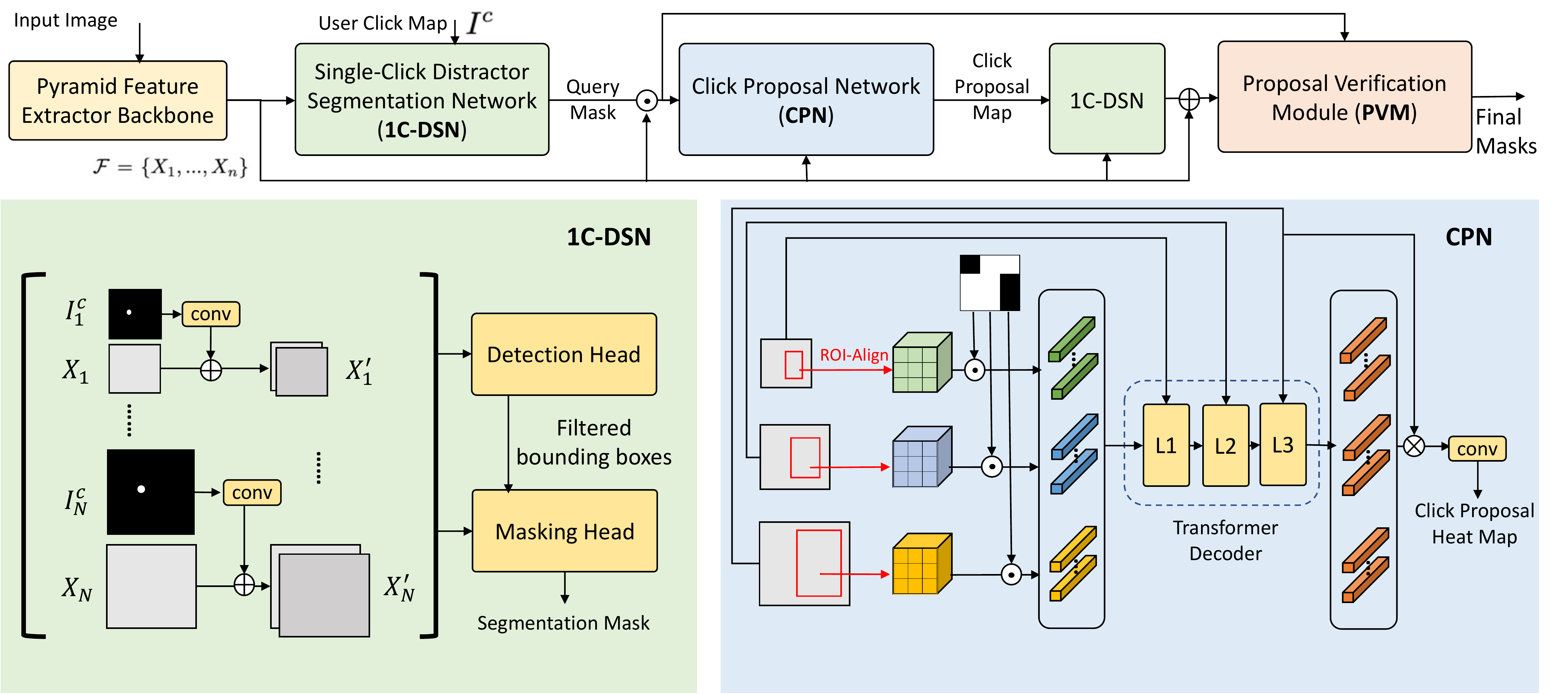}
\caption{The overview of SimpSON framework with 1C-DSN, CPN and PVM modules. It consists of a feature extraction backbone, a single-click Distractor Segmentation Network (1C-DSN), a similarity Click Proposal Network (CPN) designed for mining all the similar clicks, and a Proposal Verification Module (PVM) to check the similarity of the proposed click positions. The process of finding similar distractors can be run iteratively to fully generate the masks.}
\vspace{-5mm}
\label{fig:pipeline}
\end{figure*}

Figure \ref{fig:pipeline} shows the overall pipeline of the proposed SimpSON pipeline. It consists of a feature extraction backbone, a single-click Distractor Segmentation Network (1C-DSN), a similarity Click Proposal Network (CPN) designed for mining all the similar clicks, and a Proposal Verification Module (PVM) to check the similarity of the proposed click positions. The process can be run iteratively. 

\subsection{One-click Distractor Segmentation Network (1C-DSN)}\label{sec:segmentation}
\label{sec:one_click}
\paragraph{Motivation.} When it comes to visual distractors in users' photos, they often come in all shapes and sizes with different appearances. We don't always know what these objects are, or how big or small they might be. To tackle this challenge, we need an interactive segmentation model that is highly adaptive, especially when dealing with unfamiliar classes or small and medium-sized objects. It should be able to respond to clicks at any position, even if they fall on rare or unexpected objects, like cigarette butts, puddles, or bird droppings on the ground. To achieve this, we need to ensure that our model is optimized for high recall, so that users can remove unwanted objects with just one click. 

\paragraph{Difference with Previous Interactive Segmentation.} When designing our pipeline, we imagined that users might wish to remove many distracting elements. For that scenario, we found it more intuitive and efficient to use only positive clicks in an iterative removal workflow, which could be particularly suited for mobile apps. As discussed in section \ref{sec:iteractive}, recent interactive segmentation works are designed for precise object segmentation with multiple positive and negative clicks. We found state-of-the-art tools like \cite{chen2022focalclick, sofiiuk2022reviving} are not friendly to small and medium object segmentation with only a few positive clicks. However, for distractor selection tasks, many objects of small size should be easier to choose with one click. Larger and medium distractors had better be quickly selected with few positive clicks. So the major difference between our segmentation model and previous works is we do not use negative clicks and fully optimize our models with fewer positive clicks. 

\paragraph{Network Structure.}
Figure \ref{fig:pipeline} shows the network structure of the single-click distractor segmentation network. Given an image $I \in \R^{H \x W \x 3}$, the feature extractor network provides a pyramid feature map: $\mathcal{F} = \{X_1, ..., X_N\}$ with $X_i \in \R ^ {h^i \x w^i \x d}$ and $H > h^1 > ... > h^N, W > w^1 > ... > w^N$. For each feature level, we pair it with a binary click map $I_i^c \in \{0,1\}^{h^i \x w^i}$ where ${I_i^c}_{x,y}=1$ indicates the click at spatial location $(x, y)$ in $I_i^c$. The click-embedded feature map $X'_i \in \R ^ {h^i \x w^i \x (d + c)}$ is then computed as $X'_i = X_i \oplus conv_i(I_i^c)$, where $\oplus$ indicates the concatenation along the feature dimension and $conv_i$ is a mapping function which projects $I_i^c$ to $\R^{h^i \x w^i \x c}$. 

After obtaining the groups of click-embedded feature map $X'_i$, we feed them to the detection head and segmentation head. We modify the bounding box filtering strategy by considering only keeping the boxes that overlap with the click positions. In this paper, we follow Entity Segmentation \cite{qi2021open} to design the detection and segmentation heads. The segmentation module finally outputs multiple binary segmentation masks $M_j \in \{0,1\}^{H \x W}$ corresponding to the user click positions. The 1C-DSN is trained with similar loss functions as in Entity Segmentation, which combines detection loss from FCOS \cite{tian2019fcos} and the DICE loss from Entity Segmentation \cite{qi2021open}. The design of the detection and segmentation parts can be replaced with any two-stage segmentation frameworks \cite{he2017mask}. 


\subsection{Click Proposal Network (CPN)}
In situations where there is only one instance of a distractor, the 1C-DSN model can be sufficient for accurately segmenting it out. However, in many cases, we may come across multiple instances of distractors that share similar categories and appearances. In such scenarios, users would prefer to be able to select all of these instances with just a single click. To address this, we have designed a self-similarity mining module that can effectively identify all the distractors that are similar to the user's click, thus enabling them to remove them all in one go.

We propose this Click Proposal Network (CPN) to mine similar regions using cross-scale feature matching and regress the click positions from the high-confident regions. Then we can feed those click coordinates back to our 1C-DSN for masking to obtain the masks of all the similar distractors. The design of the Click Proposal Network (CPN) is shown in Figure \ref{fig:pipeline}. The input to the CPN is a single query mask predicted from the previous 1C-DSN corresponding to the user's single click. We utilize three levels of feature maps with the spatial resolution to be $\frac{1}{4}$, $\frac{1}{8}$, and $\frac{1}{16}$ of the input image size. For the given query mask region, we apply ROI-Align \cite{he2017mask} to extract features from the three levels of maps, resize them to $k\times k\times d$, where $k=3$ is a hyper-parameter for query size and $d$ is the dimension of the features, and then apply the binary query mask to zero-out non-masking feature regions. We then obtain $3 \times k^2$ feature vectors for similarity comparison with the original feature maps. We feed the query vectors into a cascade of transformer decoder layers L1, L2, and L3, where each layer takes the keys and values from different levels of feature maps. We finally use the obtained aggregated feature vector to conduct spatial convolution with the largest feature map to obtain the prediction click position heatmap. 

During training, we follow CenterNet \cite{zhou2019objects} to generate the ground truth heatmap using Gaussian filtering of the click map. The kernel size of the gaussian filter is set to the minimum value of the height and width of each mask. The module is then trained using a penalty-reduced pixel-wise logistic regression with focal loss as in CenterNet. During inference, we apply the Non-Maximum Suppression (NMS) to the heatmap to keep only the maximum value within a $s\times s$ window and choose all the clicks having confidence larger than $\tau_c$. Empirically, we set $s=32$ and $\tau_c=0.2$.

\subsection{Proposal Verification Module (PVM)}
\begin{figure}[t]
\centering
\captionsetup{type=figure}
\includegraphics[width=1.\linewidth]{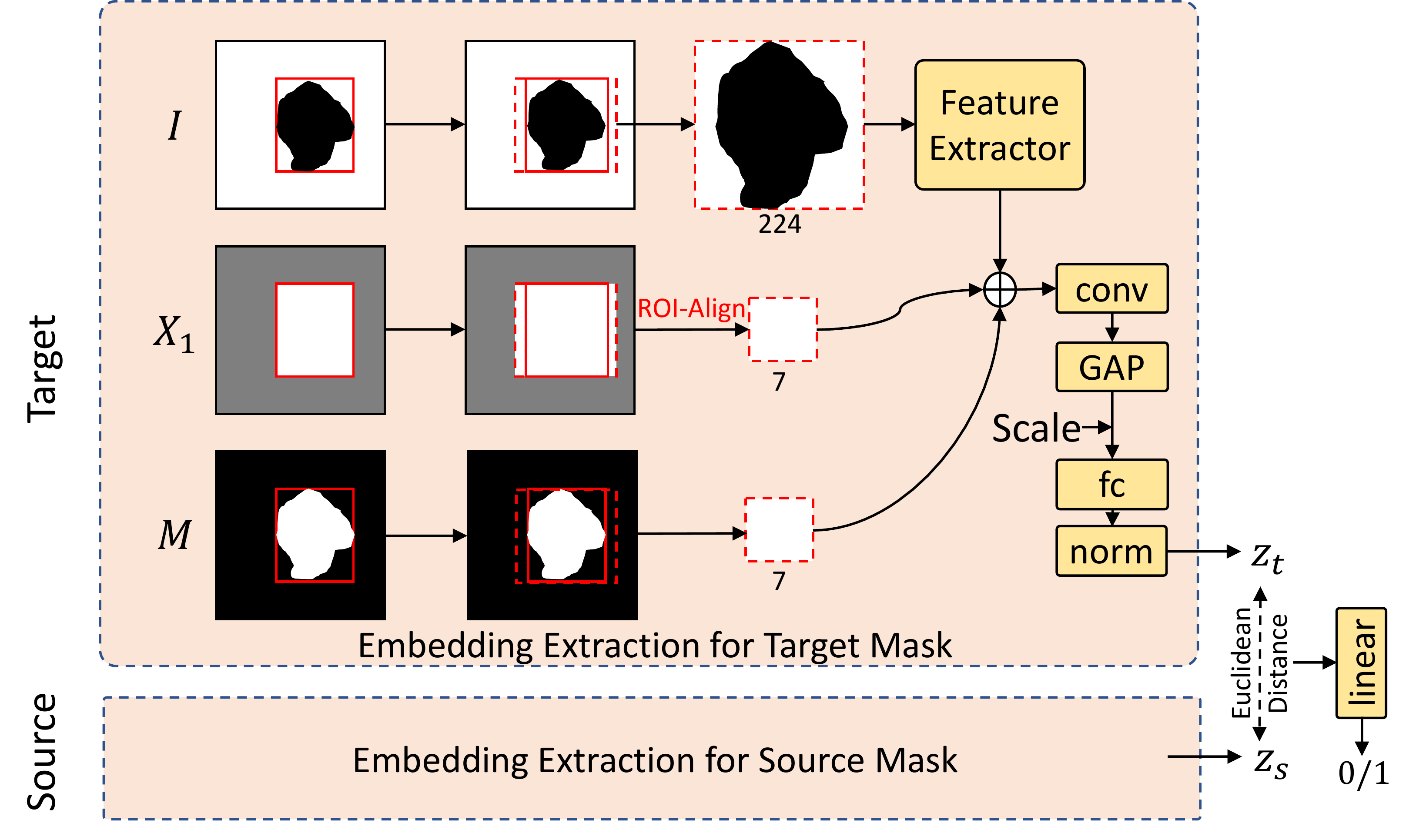}
\vspace{-6mm}
\caption{Proposal Verification Module (PVM). Given the original image $I$, the features $X_1$, and the segmentation mask $M$, we extract the region of interests from them. We then concatenate and feed the features from $I$, $X_1$ and $M$ to obtain the 1D feature embedding, $z_t$ for the target and $z_s$ for the source. The Euclidean distance between them is fed to the fully-connected layer with a sigmoid activation to output the similarity score from 0 to 1.}
\vspace{-5mm}
\label{fig:pvm}
\end{figure}

To avoid false positive proposals in the heatmap and click map, we propose using a Proposal Verification Module (PVM) to ensure that the selected click positions are highly similar to the user's clicks. This module performs pairwise comparisons between the generated masks and the initial click, and removes any click proposals that generate a mask that is significantly different from the initial query mask using a threshold.

Specifically, we first feed all the click proposals into the 1C-DSN to generate separate instance masks for each click position. We refer to the mask of the initial user click as the target mask and all the other proposed masks as source masks. Figure \ref{fig:pvm} shows the module structure of PVM and the process of comparing two distractors. Given the original image $I$, the features $X_1$, which is $\frac{1}{4}$ of the spatial image resolution, extracted from the pre-trained feature backbone in the 1C-DSN, and the segmentation mask $M$, we extract the region of interests from them. To preserve the aspect ratio of the objects, we extend the bounding box to square and use ROI-Align \cite{he2017mask} to extract pixels or features. In this paper, we resize the cropped image patch to $224 \times 224$ and feed it into a lightweight feature extractor, ResNet18 \cite{he2016deep}. We then concatenate the image features (from $I$), backbone features (from $X_1$), and resized masks (from $M$) together and feed them into neural layers to obtain the 1D feature embeddings, $z_t$ for the target and $z_s$ for the source. Notice that we also add the scaling factor $\frac{w_b}{224}$ to guide the embedding learning, where $w_b$ is the bounding box size. The Euclidean distance between $z_s$ and $z_t$ is input to the next fully-connected layer with a sigmoid activation to output the similarity score from 0 to 1.

In training, we randomly sample pairs from the same image. A pair is considered positive if it is drawn from the same copy; otherwise, it will be a negative pair. Besides the binary cross entropy $\mathcal{L}_{BCE}$ is computed on the last output with the pair labels, the max-margin contrastive loss~\cite{max_margin_contrloss} $\mathcal{L}_{con}$ is integrated on feature embedding $z_t, z_s$ to make the model learning features better. The final training loss is a linear combination $\mathcal{L} = \mathcal{L}_{con} + \mathcal{L}_{BCE}$. In testing, the PVM classifies each mask proposal with its exemplar by thresholding the similarity score. In our experiments, we choose 0.5 as the threshold.


\subsection{Iterative Distractor Selection (IDS)}
We further run an iterative process to sample more similar distractors to ensure that we entirely select all the distractors similar to the initial click. The details pseudo-code is shown in Algorithm \ref{alg:ids}. We update the $M_e$ with the correct masks for each iteration and progressively add high-confidence clicks to the result. By updating $M_e$, we can avoid incorrect similarity findings caused by the incomplete initial exemplar mask. Picking top-$k$ clicks and PVM module is essential in reducing false positive rates of CPN. In our experiments, we choose a kernel size of 5 for NMS, $N=5$, $k=10$, and $m=3$.

\begin{algorithm}
\caption{IDS: Iterative Distractor Selection}\label{alg:ids}
\KwData{$M_{init}$ (Initial Mask), $M_e$ (Examplar Set), $M_{acc}$ (Accepted Masks), $C_{acc}$ (Accepted Clicks), $N$ (maximum iteractions)}
\KwResult{$M_{acc}$, $C_{acc}$}
$itr \gets 0$\;
$M_e=M_{init}$\; 
$M_{acc} \gets \{M_{init}\}$\;
$C_{acc} \gets \emptyset$\;
\While{$itr \leq N$}{
  Generate Heatmap Using $M_e$ in CPN\;
  Apply NMS to obtain clicks $C_{new}$\;
  Remove Clicks from $C_{new}$ if within $M_{acc}$\;
  $C'_{new} \gets$ top-$k$ clicks with confidence $\geq 0.2$\;
  $C_{acc} \gets C_{acc} + C'_{new}$\;
  Pass $C_{acc}$ to 1C-DSN and Run PVM for $M_{new}$\;
  $M_{acc} \gets M_{new}$\;
  $M_e \gets$ top-$m$ confident masks\;
}
\end{algorithm}

%% file: sections/4_experiments.tex
\section{Dataset Preparation}\label{sec:data}
\paragraph{Public Datasets}
We conducted single-click segmentation experiments on the public COCO Panoptic and LVIS datasets. We pre-trained our model on the COCO Panoptic dataset, which contains 118,287 images, and fine-tuned it on the LVIS dataset, which contains 1,270,141 objects across 99,388 images. Since there is some overlap between the LVIS validation set and the COCO train set, we only used 4,809 images with 50,672 instances from the original LVIS validation set for our evaluation.

\paragraph{Self-Collected Distractor Datasets}

To gain a better understanding of the distractors in users' photos and improve the quality of our masking, we curated and annotated a dataset of distractor images. We began by creating a list of common distractors found in photos, such as distracting people, shadows, lens flare, cigarette butts on the floor, construction cones, and so on. We then collected images from various public image websites, including but not limited to Flickr, Unsplash, and Pixabay, among others. To annotate our dataset of distractor images, we recruited three professional photographers to manually select and mask the distracting regions in each image that affect its overall aesthetic appeal. We found that having three annotators was sufficient to label all the distractors in a given photo. In total, our dataset contains 21,821 images, of which we used 20,790 images containing 178,815 distracting instances for training, and 1,031 images containing 8,956 instances for validation and evaluation. We have named our distractor dataset ``Distractor20K" and the evaluation dataset ``DistractorReal-Val" in this paper.

\paragraph{Data Synthesis for Similar Distractors Mining} 
During the process of collecting our dataset, we observed that it is quite common for small, similar distractors (like bird droppings on the ground) to coexist in a single photo. However, our annotators may not be able to completely mask them. To our knowledge, there is no public dataset that includes annotations for these repeated distractors that we could use to train and evaluate our CPN model. Therefore, we propose a procedure to synthesize and generate similar distractors. This approach is inspired by \cite{copypaste}, which demonstrated that copy-pasting can be an effective data augmentation technique for instance segmentation tasks.

To synthesize additional distractor data for our ``Distractor20K" dataset, we utilized instances from the LVIS dataset and adopted the Mask2Former~\cite{cheng2022masked} approach to obtain semantic segmentation masks of the images. We only synthesized distractors within the same semantic regions, including ground, ceiling, wall, sky, sea, and river, as candidate regions. We first chose to copy objects that were either existing annotated distractors within those candidate regions or from the LVIS dataset. The LVIS examples were added to ensure a minimum of three objects to copy for each region, and the ratio between the objects and semantic regions determined the number of copies with a maximum of 10. We then iteratively placed the object at the maximum position in the distance map of the semantic region and recomputed the distance map after each iteration. In total, we obtained ``DistractorSyn14K" with 14,264 images and 287,150 instances, which were used to train the CPN and PVM modules. We also created an evaluation dataset of 531 images, which we named ``DistractorSyn-Val," containing 1,188 images with 10,980 instances.



\section{Experiments}
\subsection{Implementation details}
\paragraph{1C-DSN Training}
Our 1C-DSN follows the structure of Entity Segmentation \cite{qi2021open}. Entity Segmentation followed FCOS \cite{tian2019fcos} to utilize P3 to P7 in the feature pyramid for detection and kernel prediction and used P2 for masking. Here $P_i$ denotes the features having $\frac{1}{2^i}$ of the spatial resolution of the input image. In our work, we intended to detect and find more medium and small distractors, so we utilized P2 to P5 features for both detection and segmentation. As described in section \ref{sec:segmentation}, we concatenate additional channels from the click map to the pyramid feature, and the channel number is 32. During training, we initialized the model from Entity Segmentation pre trained on COCO Panoptic Dataset \cite{coco}, and finetuned it on LVIS dataset \cite{lvis} in 4 epochs. For a better masking quality on distractor-like objects, after we obtained the model trained on LVIS, we also finetuned it on our Distractor20K dataset in 12 epochs. Our model was evaluated on both the LVIS validation set and the DistractorReal-Val dataset. 

We randomly selected at most 50\% of the instances during training to reduce the false positive rate and make the prediction results better correlated with the input click positions. For each instance, we randomly sampled 1-5 clicks by computing the distance transform and randomly putting the click around the center of the objects. 

\vspace{-1em}
\paragraph{CPN and PVM Training}
The CPN and PVM were trained on our synthetic distractor dataset containing many groups of similar distractors within one single image. To preserve the masking quality and avoid it from being affected by the fake masks and learning from composition artifacts, we freeze the 1C-DSN network and the backbones and reuse the learned feature pyramid. In CPN, we reused the features P2 to P4. In PVM, we only used P2 for feature extraction. When training the CPN, we randomly picked the target click, and the ground truth will be the groups of instances similar to the target object. While training the PVM, we randomly selected pairs of instances within the same image and assigned the labels according to their group identity. We constantly utilized 1C-DSN to generate masks for CPN and PVM during training. Both modules are trained in 14 epochs with an initial learning rate of $0.0001$ for CPN and $0.005$ for PVM, decreasing ten times at epochs 11 and 13. They are also trained with 8 A100 GPUs with batch size 16.

\subsection{Evaluation on 1C-DSN}
\paragraph{Click Embedding}  

\begin{figure}[t]
    \centering
    \vspace{-1em}
    \begin{subfigure}[b]{0.49\columnwidth}
        \centering
        \includegraphics[width=\columnwidth]{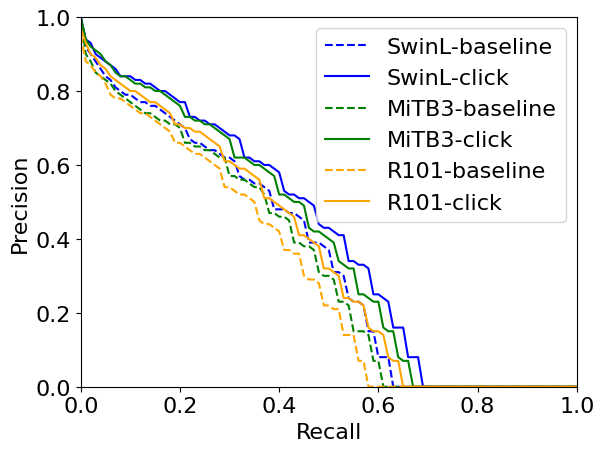}
        \caption{PR on LVIS Validation Set.}
    \end{subfigure}
    \begin{subfigure}[b]{0.49\columnwidth}
        \centering
        \includegraphics[width=\columnwidth]{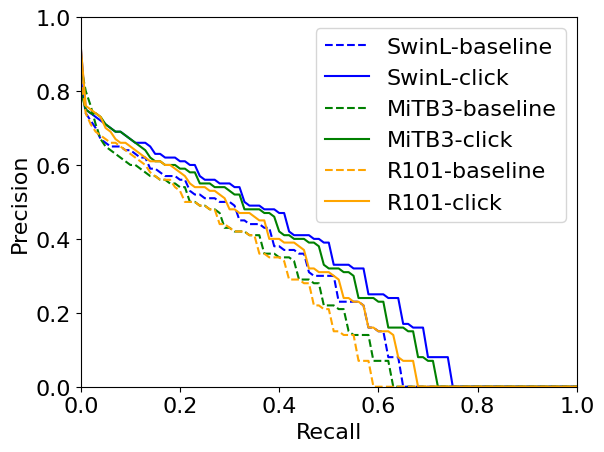}
        \caption{PR on DistractorReal-Val.}
    \end{subfigure}
    \caption{Precision-Recall (PR) curve on the validation dataset comparing the baseline and our proposed single-click based segmentation. }
    \label{fig:PR}
\end{figure}

\begin{table}[t]
    \centering
    \footnotesize 
    \setlength{\tabcolsep}{1pt} 
    \vspace{-1em}
    \begin{tabu} to \columnwidth {@{}X[3,l]X[4,c]X[2,c]X[2,c]X[2,c]X[2,c]@{}}
    \toprule
    Backbone & Click Embedding & AP & $\text{AP}_s$ & $\text{AP}_m$ & $\text{AP}_l$ \\
    \midrule
    R101 & & 30.7 & 20.6 & 47.0 & 27.9 \\
    R101 & \checkmark & 35.5 & 25.8 & 53.1 & 31.5 \\
    \midrule
    MiT-B3 & & 33.2 & 22.5 & 50.6 & 30.2\\
    MiT-B3 & \checkmark & 38.5 & 27.8 & 57.1 & 35.3\\
    \midrule
    Swin-L &  & 35.1 & 24.9 & 53.9 & 30.2 \\
    Swin-L & \checkmark & 40.2 & 31.1 & 59.0 & 35.1 \\
    \bottomrule
    \end{tabu}
    \caption{Single-click segmentation on LVIS validation set. All models are pretrained on COCO Panoptic 2017 dataset.}
    \label{tab:lvis_oneclick}
    \vspace{-1em}
\end{table}

\begin{table}[t]
    \centering
    \footnotesize 
    \setlength{\tabcolsep}{1pt} 
    \begin{tabu} to \columnwidth {@{}X[3,l]X[4,c]X[2,c]X[2,c]X[2,c]X[2,c]@{}}
    \toprule
    Backbone & Click Embedding & AP & $\text{AP}_s$ & $\text{AP}_m$ & $\text{AP}_l$ \\
    \midrule
    R101 & & 25.2 & 18.3 & 34.4 & 28.1 \\
    R101 & \checkmark & 29.9 & 23.5 & 39.2 & 32.7 \\
    \midrule
    MiT-B3 & & 26.2 & 18.6 & 35.9 & 28.3\\
    MiT-B3 & \checkmark & 32.2 & 25.1 & 43.3 & 35.9 \\
    \midrule
    Swin-L &  & 28.5 & 23.0 & 36.3 & 32.5 \\
    Swin-L & \checkmark  & 34.0 & 28.2 & 41.9 & 38.0 \\
    \bottomrule
    \end{tabu}
    \caption{Single-click segmentation on Distractor validation set. The click-embedding module outperforms 4.9 AP with R101 backbone and 8.3 with Swin-L. All models are pretrained on LVIS dataset.}
    \label{tab:distractor_oneclick}
    \vspace{-1em}
\end{table}

To evaluate the importance of click embedding for improving the performance, especially the recall rate of the model, we compared it with a baseline that was trained without click embedding as the input. We use the same click positions when comparing them. But for the baseline, we used the click positions to extract the masks which have an overlap with the clicks for evaluation. Figure \ref{fig:PR} shows the Precision-Recall (PR) curve, which demonstrates that click inputs drive the segmentation process to focus on the users' click positions and improve the overall precision and recall for all the feature extraction backbones. Table \ref{tab:lvis_oneclick} and \ref{tab:distractor_oneclick} show the Average Precision (AP) while testing on the LVIS validation dataset and our DistractorReal-Val. We split our instances into small ($\leq 32 \times 32$), medium ($32 \times 32$ to $96 \times 96$), and large ($\geq 96 \times 96$) and evaluated them separately. The gains of the Average Precision (AP) show the evidence that click embedding helps improve the segmentation performance.

\paragraph{Comparisons with Interactive Segmentation.}

\begin{figure}
    \centering
    \vspace{-1em}
    \begin{subfigure}[b]{0.49\columnwidth}
        \centering
        \includegraphics[width=\columnwidth]{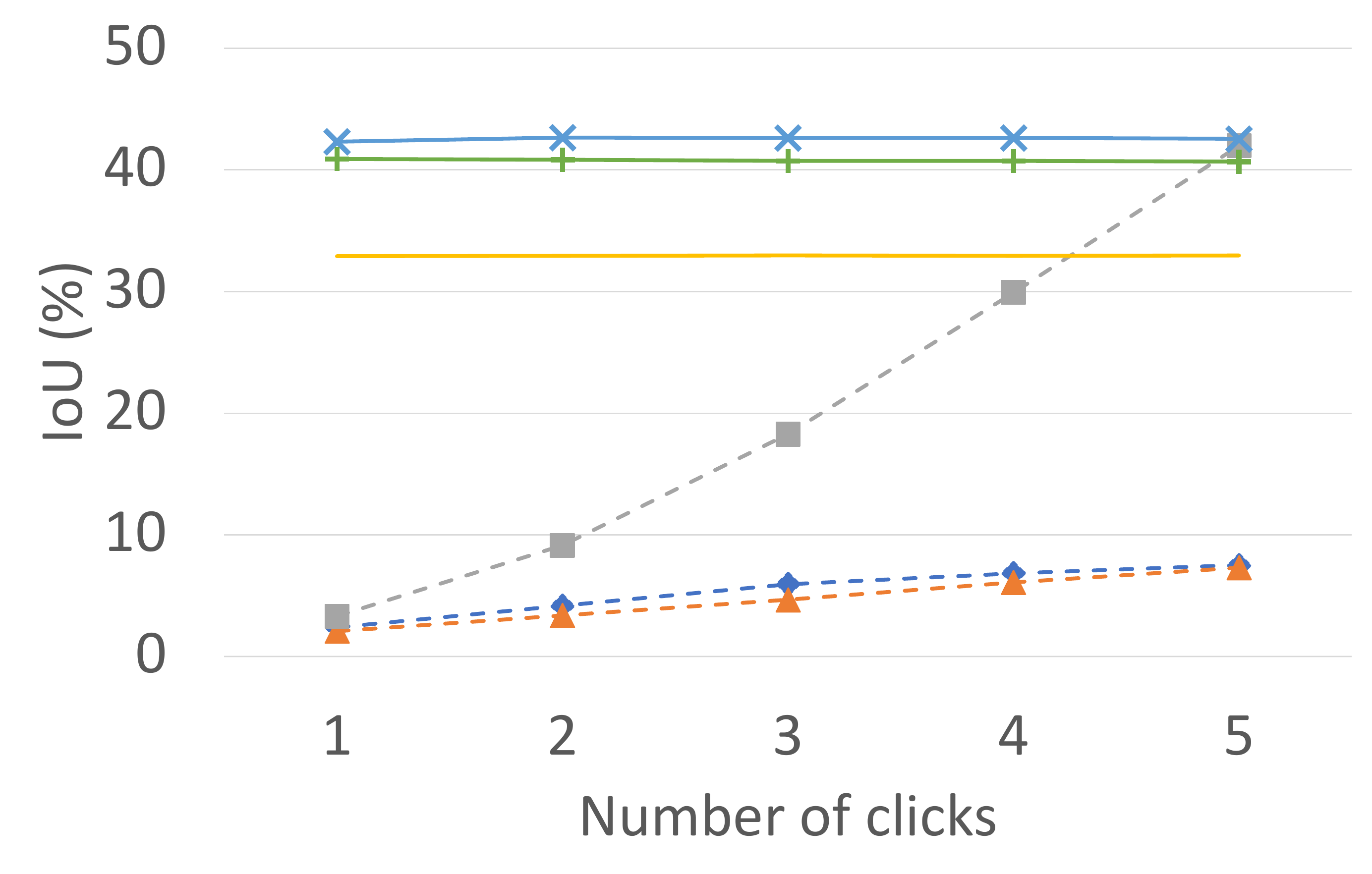}
        \caption{Small (LVIS Val)}
    \end{subfigure}
    \begin{subfigure}[b]{0.49\columnwidth}
        \centering
        \includegraphics[width=\columnwidth]{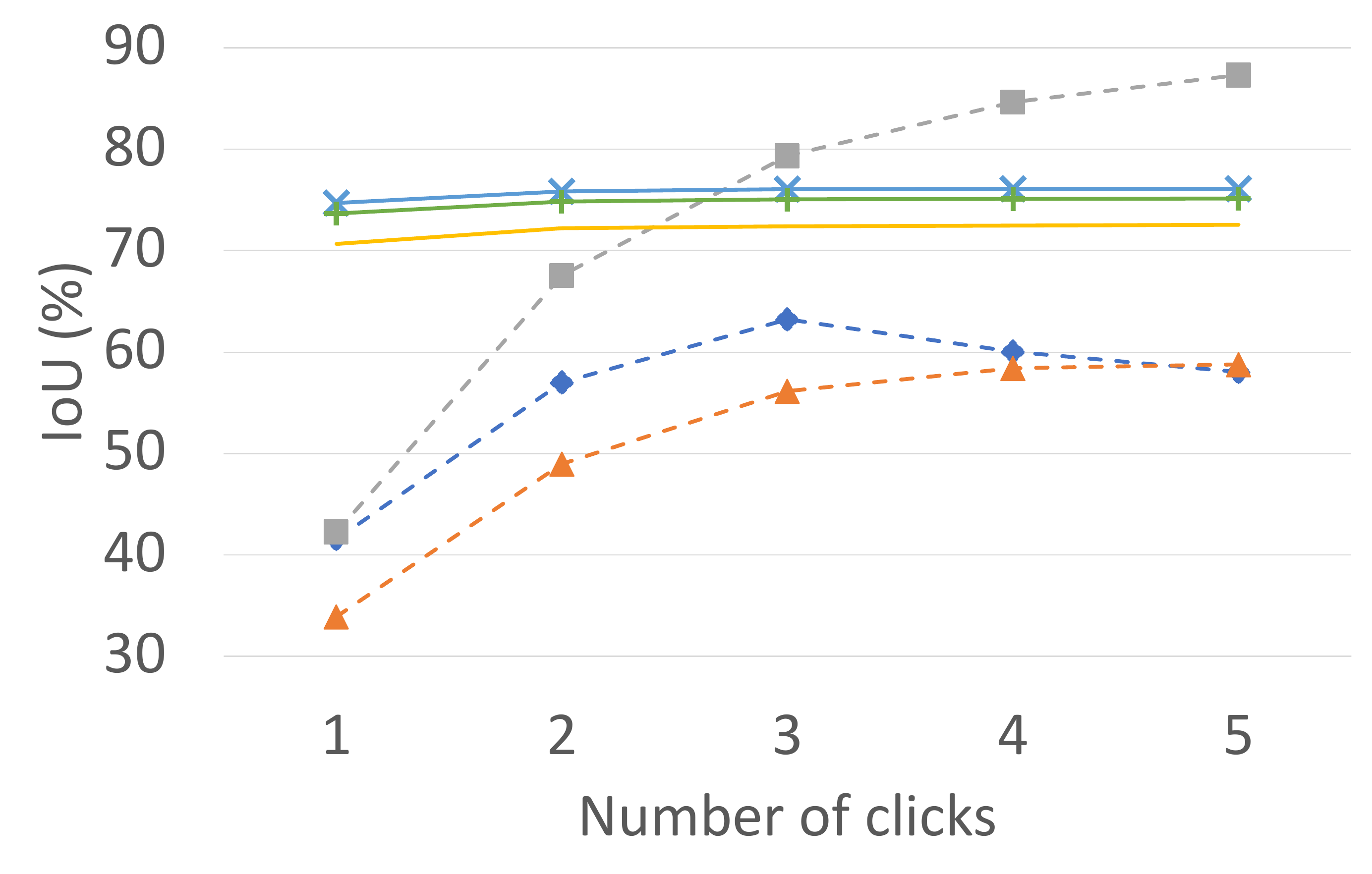}
        \caption{Medium (LVIS Val)}
    \end{subfigure}
    \begin{subfigure}[b]{0.49\columnwidth}
        \centering
        \includegraphics[width=\columnwidth]{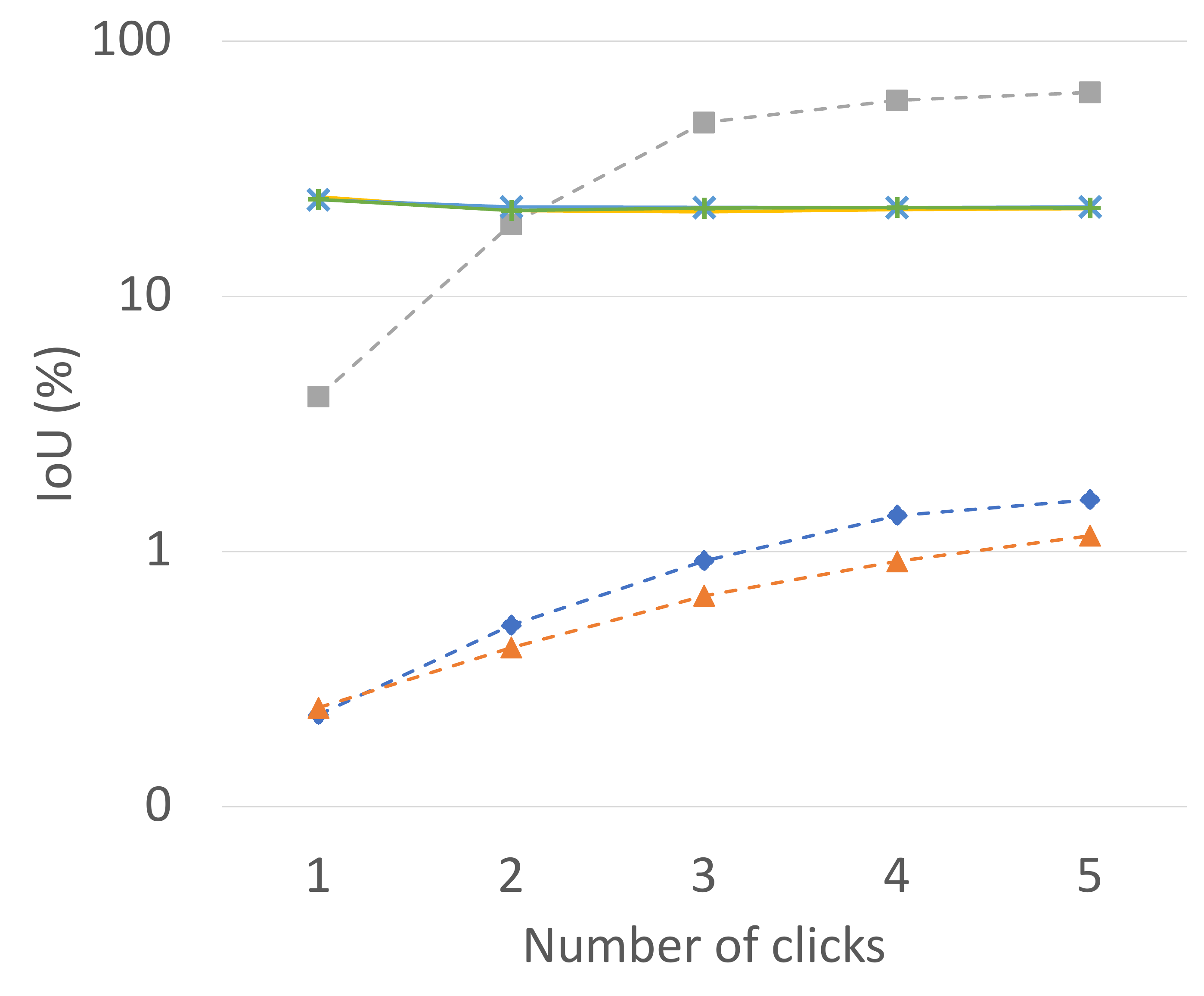}
        \caption{Small (DistractorReal-Val)}
    \end{subfigure}
    \begin{subfigure}[b]{0.49\columnwidth}
        \centering
        \includegraphics[width=\columnwidth]{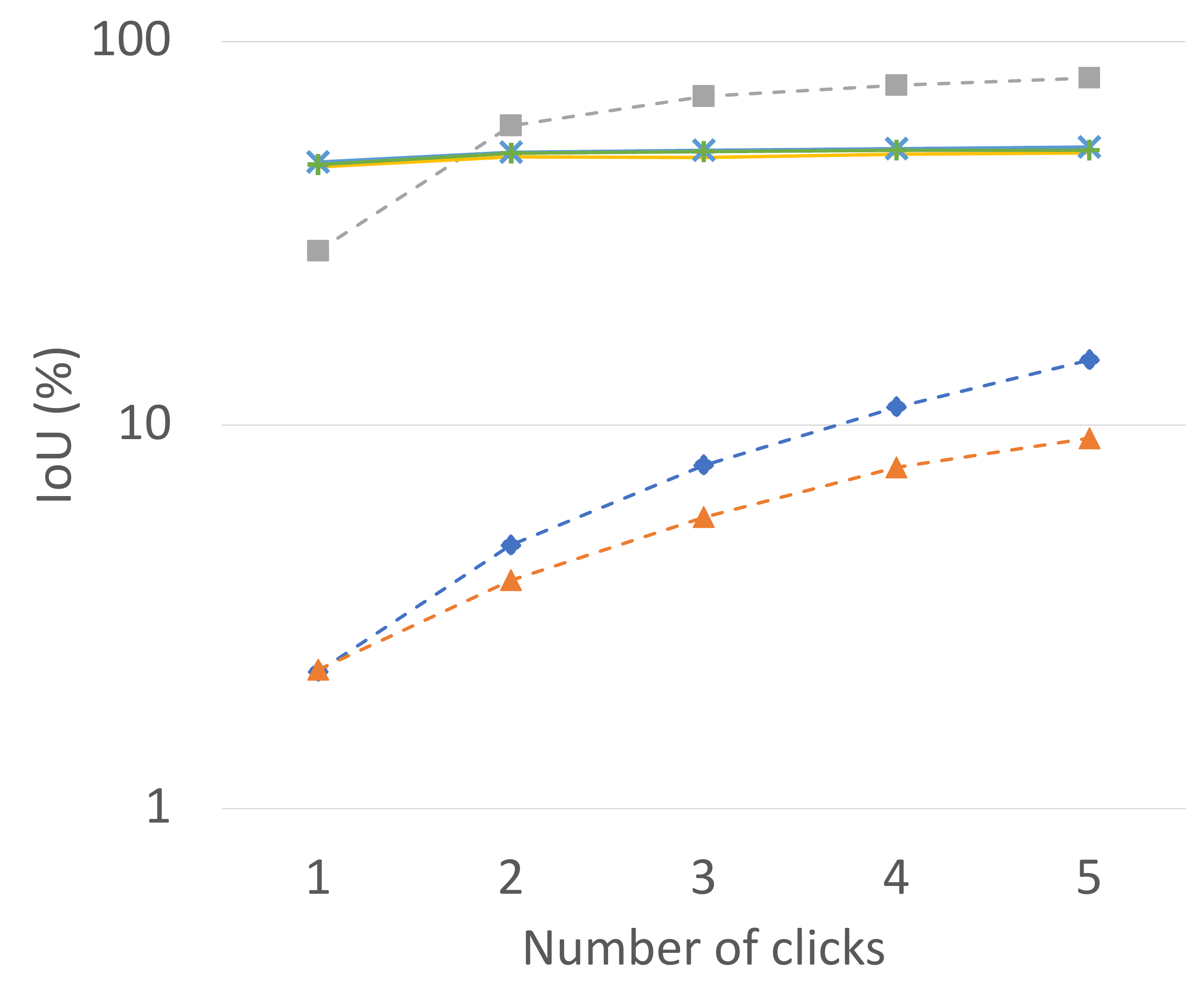}
        \caption{Medium (DistractorReal-Val)}
    \end{subfigure}
    \includegraphics[width=\columnwidth]{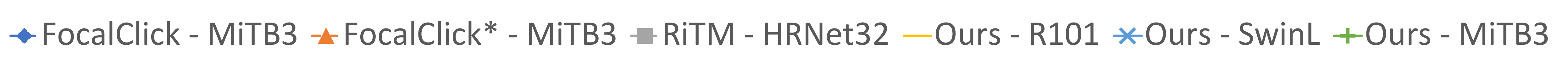}
    \caption{IoU comparison among different state-of-the-art interactive segmentation works including RiTM \cite{sofiiuk2022reviving} and FocalClick \cite{chen2022focalclick} on LVIS validation set and our DistractorReal-Val set. 
    }
    \label{fig:clickno} 
\end{figure}

Though our method is trained with only positive clicks for the specific distractor removal applications, it is still worth comparing our model with other state-of-the-art interactive segmentation in terms of precision and interaction behaviors. In this paper, we compared RiTM \cite{sofiiuk2022reviving}, and FocalClick \cite{chen2022focalclick} by using their optimal iterative strategies of sampling positive and negative clicks during testing to mimic user interaction behaviors. For our method, we follow RiTM to form the positive click sequence. Since we do not have negative clicks, we only check the False Negative (FN) region for the new next click and compute the peak value of the distance map of the FN region to place the click. Figure \ref{fig:clickno} shows the changes in average IoU as we added more clicks. For a fair comparison, all the models were trained using the same combined COCO and LVIS dataset, and some baselines have the same feature backbones. We tested them on both the LVIS validation set and our DistractorReal-Val set. 

As shown in Figure \ref{fig:clickno}, our model, regardless of feature backbones, had a steady behavior for small and medium object segmentation. Notice that we achieved a high IoU with only one single click. The behavior is desirable for distractor selection since users may want a quick and precise selection for small and medium objects but not use more clicks to refine the boundary. The curves of RiTM and FocalClick are both increasing due to the existence of negative clicks, so their methods can remove false positive regions to improve the segmentation in the process. However, relying more on negative clicks during training also worsens the first-click results. For the distractor selection task, our method has two advantages: high response to the first click and keeping steady and better while adding more clicks without causing large false positive regions. More results are in the supplementary materials.

\begin{table}[t!]
    \centering
    \footnotesize 
    \setlength{\tabcolsep}{1pt} 
    \begin{tabu} to \columnwidth {X[1.2, c]X[1.2, r]X[1, c]X[1, c]X[1, c]X[1, c]X[1, c]X[1, c]X[1, c]X[1, c]}
     \toprule
     IDS & PVM & AP & $\text{AP}_s$ & $\text{AP}_m$ & $\text{AP}_l$ & AR & $\text{AR}_s$ & $\text{AR}_m$ & $\text{AR}_l$ \\
     \midrule
     & & 34.1 & 21.0 & 35.2 & 39.4 & 41.0 & 31.1 & 41.1 & 50.9 \\
     & \checkmark & 33.7 & 21.9 & 34.6 & 39.5 & 39.0 & 30.6 & 39.2 & 47.2 \\
     \checkmark & & 34.4 & 18.7 & 35.8 & 42.5 & 47.0 & 36.3 & 47.2 & \textbf{57.4} \\
     \checkmark & \checkmark & \textbf{42.4} & \textbf{35.6} & \textbf{43.4} & \textbf{44.2} & \textbf{49.7} & \textbf{44.5} & \textbf{50.5} & 54.2 \\
     \bottomrule
    \end{tabu}
    \caption{The Group Selection Performance Gain Using IDS and PVM modules. IDS: Iterative Distractor Selection, PVM: Proposal Verification Module.}
    \label{tab:group}
    \vspace{-2.0em}
\end{table}

\begin{figure*}[t]
\centering
\vspace{-1em}
\captionsetup{type=figure}
\includegraphics[width=1.\linewidth]{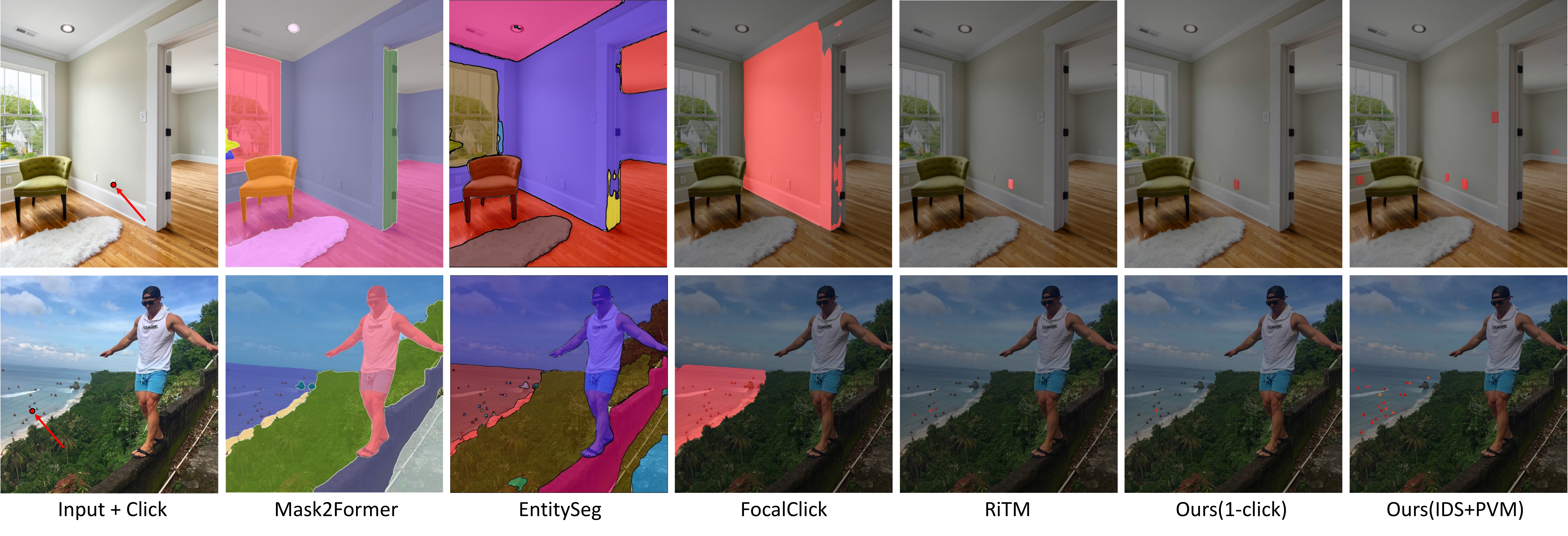}
\caption{Distractor selection comparison using different off-the-shelf segmentation models on our real user images (upper row) and synthetic data (bottom row). Models trained for panoptic segmentation tasks like Mask2Former and EntitySeg cannot focus on small and tiny objects well. Interactive segmentation works rely one negative clicks to shrink the selecting regions, and they cannot behave like clicking-one-selecting-all. Our SimpSON works well for small and tiny distractors, and can select similar things in a group. }
\label{fig:comp}
\end{figure*}

\subsection{Group Distractor Selection}


We evaluated our group selection performance on the DistractorSyn-Val dataset using the proposed CPN, PVM, and the iterative process (IDS). Table \ref{tab:group} lists the performance difference if we change the pipeline components when running the selection. Recall that after we apply CPN to propose clicks and feed those clicks to 1C-DSN for masking, we can use PVM to reject false positives, so it possibly decreases the overall recall rate a little bit. At the same time, the iterative process (IDS) will generate more click proposals in the photos to boost the recall rate. Combining the two strategies (IDS and PVM), therefore, yields the best overall performance on our synthetic validation set. Figure \ref{fig:comp} shows some examples when testing the model on both real and synthetic data. Compared with other off-the-shelf segmentation models, our single-click based model has a higher response to tiny distractors and is functional in interactive group selection. Our 1C-DSN is trained on a real distractor dataset, while the group selection pipeline is trained on a synthetic dataset. We found our model generalizes well to find similar objects in real images in Figure \ref{fig:tea} and \ref{fig:comp}. 


\subsection{More Ablation Studies}

\paragraph{Ablations on CPN and PVM Module.}
\begin{table}[t!]
    \centering
    \footnotesize 
    \setlength{\tabcolsep}{1pt} 
    \begin{tabu} to \columnwidth {@{}X[3,l]X[1,c]X[2,c]X[2.5,c]@{}}
         \toprule
         $L1 \rightarrow L2 \rightarrow L3$ & Mask & Query size & AUC-PR (\%) \\
         \midrule
         $1/4 \rightarrow 1/8 \rightarrow 1/16$ & \checkmark & $3 \x 3$ & \textbf{40.43}\\
         $1/4 \rightarrow 1/8 \rightarrow 1/16$ &  & $3 \x 3$ & 35.08\\
         $1/16 \rightarrow 1/8 \rightarrow 1/4$ & \checkmark  & $3 \x 3$ & 37.00 \\
         $1/4 \rightarrow 1/8 \rightarrow 1/16$ & \checkmark & $5 \x 5$ & 36.62 \\
         $1/4 \rightarrow 1/8 \rightarrow 1/16$ & \checkmark & $7 \x 7$ & 34.14 \\
         \bottomrule
    \end{tabu}
    \caption{Ablation study on Click Proposal Network (CPN) on DistractorSyn-Val.} 
    \label{tab:cpn}
    \vspace{-1.2em}
\end{table}

We conducted ablation studies on the design of Click Proposal Network (CPN) in Table \ref{tab:cpn}. We found that zeroing out irrelevant feature patches using masking was necessary to avoid a high false positive rate. If we enlarged the query patch size, the query vector would become more localized, so it yielded a higher false positive rate. The order of the feature map inputting to different layers of the transformer decoder was also important since starting the matching from the largest feature map would possibly lead to better feature aggregation. Several design details of the Proposal Verification Module (PVM) have been compared in Table \ref{tab:pvm}. 
Our ablation experiments demonstrate that all three designs contribute to improving precision and recall. 

\begin{table}[t!]
    \centering
    
    \scriptsize
    \begin{tabu} to \columnwidth {X[1.2, c]X[1.2, r]X[1.2, r]X[1, c]X[1, c]X[1, c]X[1, c]X[1, c]X[1, c]X[1, c]X[1, c]}
     \toprule
     Scale & Square & Mask & AP & $\text{AP}_s$ & $\text{AP}_m$ & $\text{AP}_l$ & AR & $\text{AR}_s$ & $\text{AR}_m$ & $\text{AR}_l$ \\
     \midrule

     & \checkmark & \checkmark & 42.2 & 35.2 & 43.2 & 43.9 & 48.5 & 44.2 & 49.7 & 53.8 \\
     \checkmark & & \checkmark & 42.3 & 34.4 & 43.3 & 44.1 & 48.7 & 44.1 & 50.2 & 54.1 \\
     \checkmark & \checkmark & & 42.0 & 33.5 & 43.1 & \textbf{44.8} & 43.7 & 43.7 & 49.4 & 53.6\\
    \checkmark & \checkmark & \checkmark & \textbf{42.4} & \textbf{35.6} & \textbf{43.4} & 44.2 & \textbf{49.7} & \textbf{44.5} & \textbf{50.5} & \textbf{54.2} \\
     
     \bottomrule
    \end{tabu}
    \caption{The performance of PVM with different input information on DistractorSyn-Val.} 
    \label{tab:pvm}
    \vspace{-1.5em}
\end{table}

%% file: sections/5_discussion_conclusion.tex
\section{Conclusions}
We presented SimpSON, an interactive selection network for distractor segmentation in photos. Distractors are often small, repeated, clustered, and belong to unknown categories. To address this challenge, we optimized a single-click-based segmentation network and mined all the distractors similar to the click using Click Proposal Network (CPN) for group selection. We found that applying the CPN iteratively and using an additional Proposal Verification Module (PVM) made the selection more robust by avoiding false positives. Our experiments demonstrated that active click-guided segmentation yields better precision-recall than passive retrieval of masks from a pre-computed segmentation map. We believe that our pipeline will simplify the process of photo retouching and inspire new directions for interactive segmentation research.